\documentclass[journal]{IEEEtran}

\usepackage{color}
\usepackage{url}
\usepackage[pdftex]{graphicx}
\usepackage{subfigure}
\usepackage{epstopdf}
\usepackage{cite}
\usepackage[cmex10]{amsmath}
\usepackage{enumerate}
\usepackage{multirow}
\usepackage{subfigure}
\usepackage{booktabs}
\usepackage{makecell}
\usepackage[labelfont=bf,labelsep=space]{caption}
\usepackage{float}
\usepackage{mathrsfs}

\graphicspath{{../graph/}{../png/}{graph/}{./graph/}} 
\DeclareGraphicsExtensions{.eps,.pdf,.png,.jpeg}
\begin{document}

\title{An End-to-End Attack on Text-based CAPTCHAs Based on Cycle-Consistent Generative Adversarial Network}
\author{
\IEEEauthorblockN{Chunhui Li\textsuperscript{1}, Xingshu Chen\textsuperscript{2*}, Haizhou Wang\textsuperscript{3*}, Yu Zhang\textsuperscript{4}, Peiming Wang\textsuperscript{5}} \\
\thanks{C. Li was with the College of Cybersecurity, Sichuan University, Chengdu 610065, China; X. Chen was with the College of Cybersecurity and the Cybersecurity Research Institute, Sichuan University, Chengdu 610065, China (e-mail: chenxsh@scu.edu.cn); H. Wang was with the College of Cybersecurity, Sichuan University, Chengdu 610065, China (e-mail: whzh.nc@scu.edu.cn); Y. Zhang was with the College of Art, Sichuan University, Chengdu 610065, China; P. Wang was with the College of Computer Science, Sichuan University, Chengdu 610065, China.}
}

\maketitle

\begin{abstract}
As a widely deployed security scheme, text-based CAPTCHAs have become more and more difficult to resist machine learning-based attacks. So far, many researchers have conducted attacking research on text-based CAPTCHAs deployed by different companies (such as Microsoft, Amazon, and Apple) and achieved certain results.However, most of these attacks have some shortcomings, such as poor portability of attack methods, requiring a series of data preprocessing steps, and relying on large amounts of labeled CAPTCHAs. In this paper, we propose an efficient and simple end-to-end attack method based on cycle-consistent generative adversarial networks. Compared with previous studies, our method greatly reduces the cost of data labeling. In addition, this method has high portability. It can attack common text-based CAPTCHA schemes only by modifying a few configuration parameters, which makes the attack easier. Firstly, we train CAPTCHA synthesizers based on the cycle-GAN to generate some fake samples. Basic recognizers based on the convolutional recurrent neural network are trained with the fake data. Subsequently, an active transfer learning method is employed to optimize the basic recognizer utilizing tiny amounts of labeled real-world CAPTCHA samples. Our approach efficiently cracked the CAPTCHA schemes deployed by 10 popular websites, indicating that our attack is likely very general. Additionally, we analyzed the current most popular anti-recognition mechanisms. The results show that the combination of more anti-recognition mechanisms can improve the security of CAPTCHA, but the improvement is limited. Conversely, generating more complex CAPTCHAs may cost more resources and reduce the availability of CAPTCHAs. 
\end{abstract}

\IEEEpeerreviewmaketitle


%
\section{Introduction}
\IEEEPARstart{W}{ith} the rapid development of the Internet, more and more online services and website resources are threatened by malicious bots. These bots perform a series of malicious actions on the Internet by simulating the identity of real users, including sending spam, stealing website resources through crawlers, etc\cite{torky2016securing,kim2019search}. Therefore, researchers have proposed a CAPTCHA (Completely Automated Public Turing test to tell Computers and Humans Apart) mechanism to generate a test for the computer to confirm whether the remote user is human automatically\cite{gelernter2016tell}. Nowadays, various CAPTCHA schemes have appeared in the industry, such as image-based CAPTCHA schemes \cite{tang2018research} and voice-based CAPTCHA schemes \cite{Shah2018Hitting}. Nonetheless, there are still many companies using text-based CAPTCHA schemes as the primary means of security and soft authentication, including Baidu, Tencent, Amazon, eBay, Microsoft, etc. One of the main reasons is that the text-based CAPTCHAs have lower development difficulty and lower maintenance cost than other types of CAPTCHA schemes. Nevertheless, the widespread use of text-based CAPTCHA schemes also brings relatively high-security risks.

\par The development of CAPTCHA technology is an adversarial process. To improve the security, researchers have tried to add more anti-recognition security mechanisms for the existing text-based CAPTCHA schemes. These popular security mechanisms include the use of multiple font styles, complicated background interference, character rotation, overlapping, distortion, two-layer structures, noisy arcs and complicated background interference, etc\cite{gao2013robustness,gao2017research,tang2018research,ye2018yet}. These security mechanisms make character segmentation difficult and make the methods based on deep learning need a large volume of training data\cite{george2017generative}. And in general, tens of thousands of samples need to be labeled to achieve high attack success rates\cite{george2017generative, goodfellow2014multidigit, zi2019end}. Furthermore, more and more websites have added many protection measures to prevent the CAPTCHAs from being downloaded maliciously by attackers, making it more challenging to gather real-world CAPTCHA samples than before.

\par In this paper, we propose a generic, efficient, and low cost end-to-end text-based CAPTCHAs breaking method. It consists of two parts: CAPTCHA synthesizers and CAPTCHA recognizers. Firstly, we utilized the CAPTCHA synthesizers based on Cycle-GAN (cycle-consistent generative adversarial network)\cite{zhu2017unpaired} to reduce the reliance on real-world data. Subsequently, we trained the CAPTCHA recognizers based on CRNN (Convolutional Recurrent Neural Network)\cite{shi2016end}. Finally, we tested our attack on text-based CAPTCHA schemes, which were deployed by the top 50 most popular websites such as Baidu, Wikipedia, Microsoft, eBay. The experimental results show that our method can not only achieve higher attack success rates but also significantly reduce the costs of labor in the attack process.

\par The main contributions of this paper can be summarized as follows:
\begin{itemize}
    \item This paper proposes a synthetic CAPTCHA generation method based on generating adversarial networks, which effectively solves the problem that it is difficult to label a large amount of training data manually. By proposing an approach based on active transfer learning, our model can learn hard samples that contains more information, which significantly reduces the demand for real-world CAPTCHA samples and the difficulty of breaking CAPTCHAs.
    \item The CAPTCHA breaking method in this paper does not require manual preprocessing of the CAPTCHAs (such as segmentation, removing interference lines and backgrounds, and restoring distortion, etc.). It is an end-to-end process, which means that the final attack results are not disturbed by the results of upstream process. And it is more effective for breaking variable-length text-based CAPTCHA schemes.
    \item We proved that text-based CAPTCHA schemes have serious security risks through comparison experiments of the existing anti-recognition mechanisms. And the CAPTCHA breaking method employed in this paper can significantly reduce the protection ability of popular websites at home and abroad against CAPTCHA breaking attacks.
\end{itemize}

\par The remainder of this paper is organized as follows: Section II briefly summarizes previous studies in breaking text-based CAPTCHAs. Section III introduces the GAN-based CAPTCHA Synthesizer and the End-to-End CAPTCHA Recognizer in detail. We evaluate the security of 10 popular real-world text-based CAPTCHA schemes and make a comprehensive analysis of the effectiveness of existing anti-recognition schemes in Section IV. Finally, Section V concludes this study. 
%
\section{Related Work}
\par Text-based CAPTCHA is one of the most widely and frequently used protection mechanisms in website systems. It usually contains several English letters or Arabic numerals. Originally, the designers of CAPTCHAs did not use an effective anti-segmentation technology, but only added a few noise points to CAPTCHA images. For this type of CAPTCHA, Mori et al.\cite{mori2003recognizing} employed shape context matching algorithms to break the EZ-Gimpy and Gimpy with a success rate of 92\% and 33\% respectively. With the development of CAPTCHA recognition technology, designers have added character distortion mechanisms to increase the difficulty of character recognition. Chellapilla et al.\cite{chellapilla2005using} recognized distorted letters by machine learning approach, and the success rate increased from 4.89\% to 66.20\%. Research studies in \cite{chellapilla2005using,simard2003using} show that the main difficulty of CAPTCHA recognition is character segmentation, and it is pointed out that the design of anti-segmentation can better guarantee the security of CAPTCHA than distorted characters. In 2008, Yan et al.\cite{yan2008low} designed a series of character segmentation strategies for the anti-segmentation mechanism-based CAPTCHA, which effectively segmented Microsoft CAPTCHA characters.

\begin{figure*} [ht]
    \centering
    \vspace{-25pt}
    \includegraphics[width=0.85\textwidth]{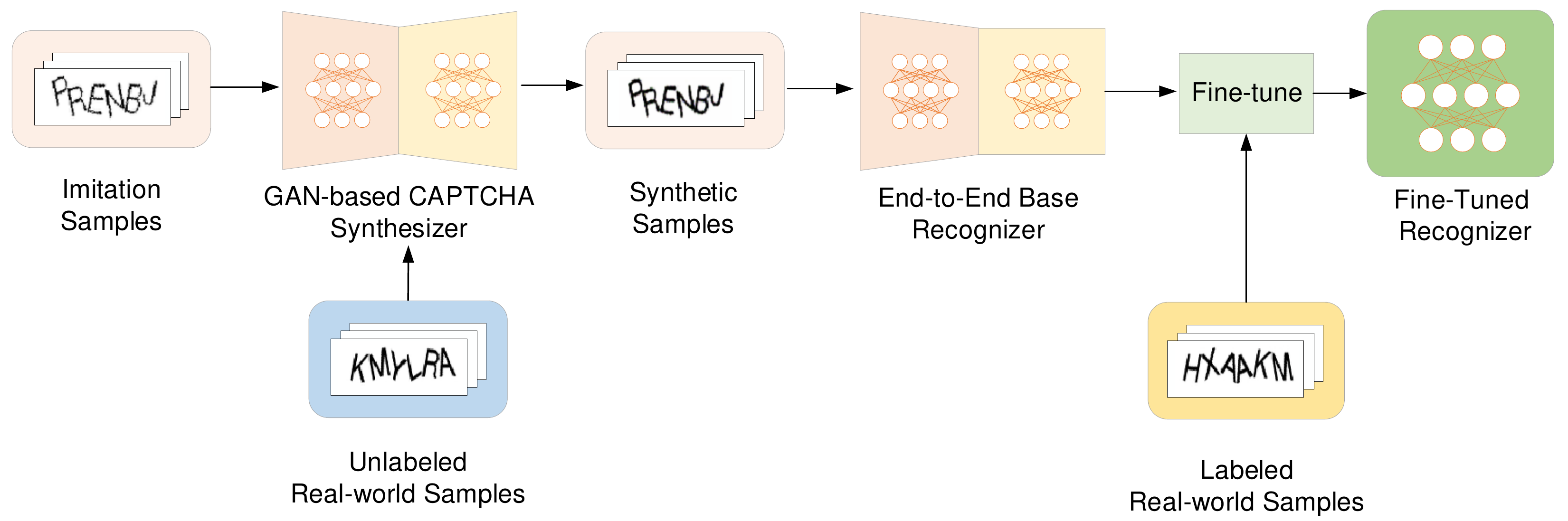}
    \caption{Overview of our CAPTCHA breaking approach. Initially, we generated some imitation CAPTCHA images. Subsequently, GAN-based CAPTCHA synthesizers were trained by using unpaired imitation samples and real-world samples. Then, The synthesizers were used to produce a great mass of synthetic samples to train the end-to-end basic recognizers. Finally, a limited number of labeled real-world CAPTCHA data were used to fine-tune the basic recognizers.}
    \label{fig:1}
    \vspace{-10pt}
\end{figure*}

\par Characters overlapping can effectively prevent the character segmentation. To tackle this problem, Franc et al.\cite{franc2005license} employed an algorithm to obtain prior knowledge of the target CAPTCHAs through HMM (Hidden Markov Chain), and then segment the images. Although this method can successfully segment overlapped characters, but it is only applicable to the same interference type CAPTCHA. Later, Starostenko et al.\cite{starostenko2015breaking} outlined a new character segmentation algorithm. This algorithm applied complex judgment conditions and prior knowledge to segment the overlapping characters and achieved a good result. Nevertheless, this algorithm fails to be generic because a lot of prior knowledge was used. Goodfellow et al.\cite{goodfellow2014multidigit} combined the positioning, segmentation, and recognition of multi-character text and built a deep CNN (Convolutional Neural Networks), then trained with millions of images containing street numbers. Finally, the success rate was over 90\%. Zhang et al.\cite{zhang2017captcha} proposed a method of CAPTCHA segmentation based on improved vertical projection, which efficiently solved the problem of different types of conglutination characters. But for italic characters, this method is not robust. Later, Tang et al.\cite{tang2018research} presented a simple, universal, and fast text-based CAPTCHA attack approach based on semantic information understanding and pixel-level segmentation. They cracked more than 13 kinds of CAPTCHA schemes that were deployed by the top 50 most popular websites in the world.

\par The success of the above attack methods depends on the effectiveness of character segmentation. With the development of CAPTCHAs, more and more sophisticated anti-recognition security mechanisms are introduced. It makes segmentation-based approach hard to work. Consequently, researchers gradually paid more attention to end-to-end methods based on deep learning. Shi et al.\cite{shi2016end} described an end-to-end image sequence recognition method based on CNN and RNN (recurrent neural networks). They applied their approach to the field of scene text recognition and achieved state of the art on multiple open test data sets at that time. Ye et al.\cite{ye2018yet} proposed a breaking method based on generating adversarial networks. They employed pre-trained models based on Pix2Pix\cite{isola2017image} to remove the anti-recognition security mechanism from the real-word CAPTCHA images and then produced synthetic CAPTCHA images without any security mechanisms. Finally, a CNN-based recognition model was trained using a limited number of target samples and a large volume of synthetic data. They had achieved impressive results from several CAPTCHA schemes. Zi et al.\cite{zi2019end} presented an end-to-end text-based CAPTCHA attack method based on convolutional neural network and attention mechanism, which can crack the target CAPTCHA without any segmentation and preprocessing. They have achieved not only good results but also experimentally showed that the existing CAPTCHA security mechanisms could not effectively resist attacks based on deep learning and they need to be strengthened urgently.

%
\section{Attack Approach}
\par In this section, we mainly introduce the GAN-based CAPTCHA synthesizer and the end-to-end CAPTCHA breaking method. Figure \ref{fig:1} shows the attack process of this paper.

\subsection{GAN-based CAPTCHA Synthesizer}
\par As discussed, the breaking approach based on machine learning and deep learning requires a large volume of labeled training data to obtain recognizers with excellent performance. However, the data labeling work is time-consuming and laborious, and requires a significant investment. With the aim to solve this problem, we trains Cycle-GAN-based CAPTCHA synthesizers. We can produce a massive number of synthetic training data by means of the CAPTCHA synthesizers to cover the real-world CAPTCHA sample spaces. This method dramatically lighten the burden of labeling data manually and significantly reduces the breaking cost.

\par \textbf{CAPTCHA Generation System}. We have designed a configurable, scalable, and flexible CAPTCHA generation system to produce unpaired data to train CAPTCHA synthesizers. The system contains some configurable settings, including font, character rotation, overlap, distortion, interference lines, etc, which almost covers the current popular anti-recognition mechanisms. We can generate a large amount of imitation CAPTCHA images that are visually similar to the real-world ones. In particular, to make the imitation CAPTCHAs similar to the real ones as much as possible, we have collected masses of fonts. These fonts are downloaded from website of fontke.com and are very similar to the target CAPTCHA scheme fonts. Meanwhile, we analyze the anti-recognition measures of the target real-world CAPTCHA schemes semi-automatically, including rotation angle, gap of two characters, and font color, etc.

\begin{table}[htb]
\centering
  \caption{THE SECURITY MECHANISMS OF WEIBO CAPTCHA SCHEME}
  \label{tab:1}
  \setlength{\tabcolsep}{0pt}  
  \renewcommand{\arraystretch}{1.3} 
    \begin{tabular}{ccc}
      \toprule[0.75pt]
    \textbf{\makecell[c]{Security \\ Mechanisms}}     & \textbf{On/Off} & \textbf{Values} \\
    \midrule[0.5pt]
    Fonts                   & On              & \makecell[c]{\{RM-Playtime-Bold.ttf,\\RM-Playtime-Medium.ttf\}} \\
    Font Colors             & On              & \{RGB(101,101, 254),RGB(254, 101, 101)\}        \\
    Excluded Characters     & On              & \{0, 1, 9, i, j, l, o, t\}                                    \\
    Rotation                & On              & {[}-10,10{]}                                                 \\
    Gaps of Characters      & On              & {[}0,2{]}                                                     \\
    Distortion              & Off             & -                                                             \\
    Lines                   & On              & \{Sin\}                                                       \\
    Background Interference & Off             & -                                                             \\ \bottomrule[0.75pt]
    \end{tabular}
    \vspace{0pt}
\end{table}

\par Taking the CAPTCHA scheme of Weibo as an example, we first downloaded two hollow fonts that are similar to the real ones from fontke.com and extracted the colors of real-world CAPTCHA fonts by a color picker to determine the font color spaces. Secondly, we ascertained the approximate range of the rotation angle range to [-10, 10] by correcting the angle of the real-world CAPTCHA characters. We settled that the excluded characters contains \{'0', '1', '9', 'i', 'j', 'l', 'o', 't'\} by counting the labels of 1,000 correctly labeled images. Then the gaps of characters were determined by calculating the number of blank pixels between two adjacent characters. In various CAPTCHA schemes, line segments, sinusoidal arcs, and Bézier arcs are commonly used as interference mechanisms \cite{ye2018yet}. We concluded that the interference line is based on a sine function \( A\ast\sin \left ( \omega x + \varphi \right ) + \sigma \) by observing a large number of Weibo samples. And this scheme draws different sine arcs by controlling parameters \( A,\omega,\varphi,\sigma \). Finally, we got the parameter values shown in Table \ref{tab:1}.

\begin{figure}[htb]
\centering
    \vspace{-5pt}
   \subfigure[]{\label{fig:2a}\includegraphics[width=0.2\textwidth]{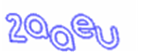}\includegraphics[width=0.2\textwidth]{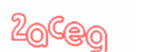}}

   \subfigure[]{\label{fig:2b}\includegraphics[width=0.2\textwidth]{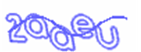}\includegraphics[width=0.2\textwidth]{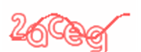}}

   \subfigure[]{\label{fig:2c}\includegraphics[width=0.2\textwidth]{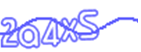}\includegraphics[width=0.2\textwidth]{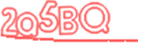}}
\caption{The comparison between the imitation samples and the real-world Weibo samples: (a) imitation samples without arc, (b) imitation samples with arc, (c) real-world samples of Weibo.}
\label{fig:2}
\vspace{-10pt}
\end{figure}

\par The imitation CAPTCHA samples generated by the system with the configuration parameters shown in Table \ref{tab:1} are compared with the real-world Weibo CAPTCHA samples, as shown in Figure \ref{fig:2}.

\par
\begin{figure} [htb]
    \centering
    \vspace{0pt}
    \includegraphics[width=0.45\textwidth]{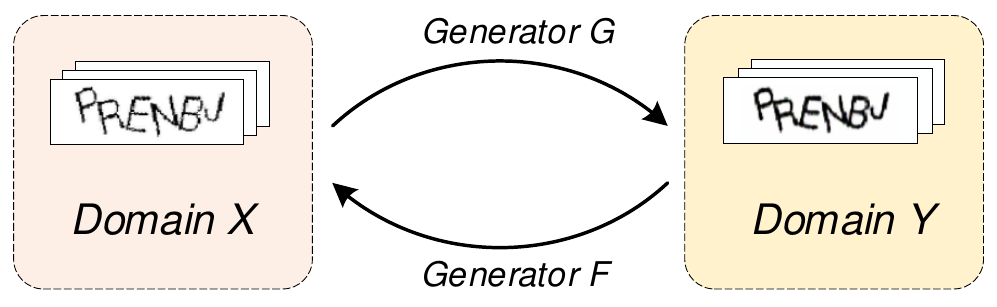}
    \caption{The overview structure of our CAPTCHA synthesizer.}
    \label{fig:3}
    \vspace{-5pt}
\end{figure}

 \textbf{CAPTCHA Synthesizer.} For the sake of reducing the gap between the imitation data and the real-world data, the imitation images need to be modified from the pixel level. That is, we need to learn a set of mappings from the given imitation images \( X=\left\{\mathop{x}_{i}\right\},i\in\left\{1,2,...,n\right\},x \sim \mathop{P}_{data}\left(x\right) \) and the real-world images \( Y=\left\{\mathop{y}_{j}\right\},j\in\left\{1,2,...,n\right\},y \sim \mathop{P}_{data}\left ( y \right ) \) to transform from the generated domain \(X\) to the target domain \(Y\) and from the target domain \(Y\) to the generated domain \(X\). Therefore, our CAPTCHA synthesizer consists of four parts: generators \(G\) and \(F\) and discriminators \(D_{X}\) and \(D_{Y}\), as shown in Figure \ref{fig:3}.

\begin{figure*}[htb]
\centering
    \vspace{-25pt}
   \subfigure[]{\label{fig:4a}\includegraphics[width=0.85\textwidth]{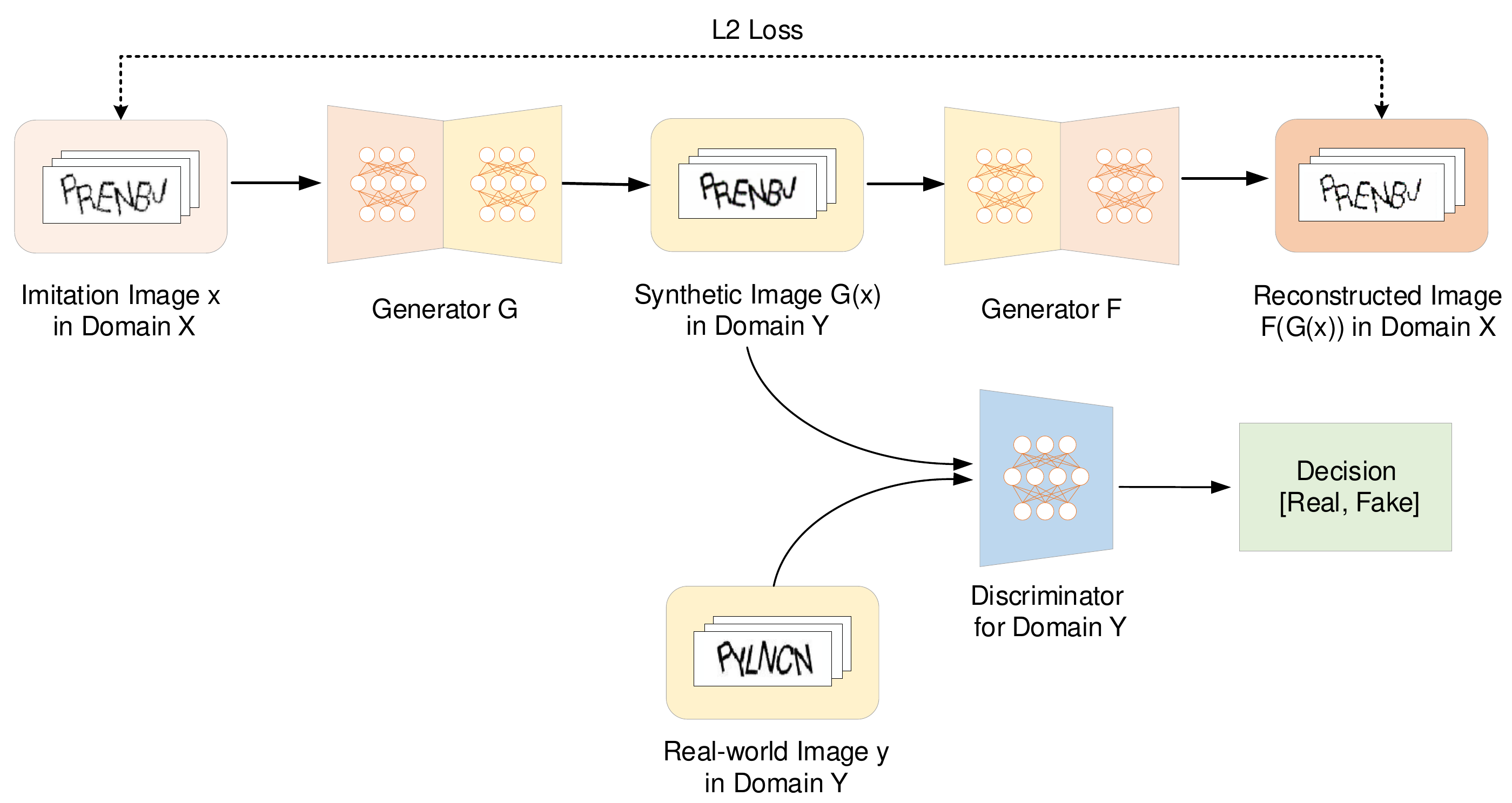}}
   \subfigure[]{\label{fig:4b}\includegraphics[width=0.45\textwidth]{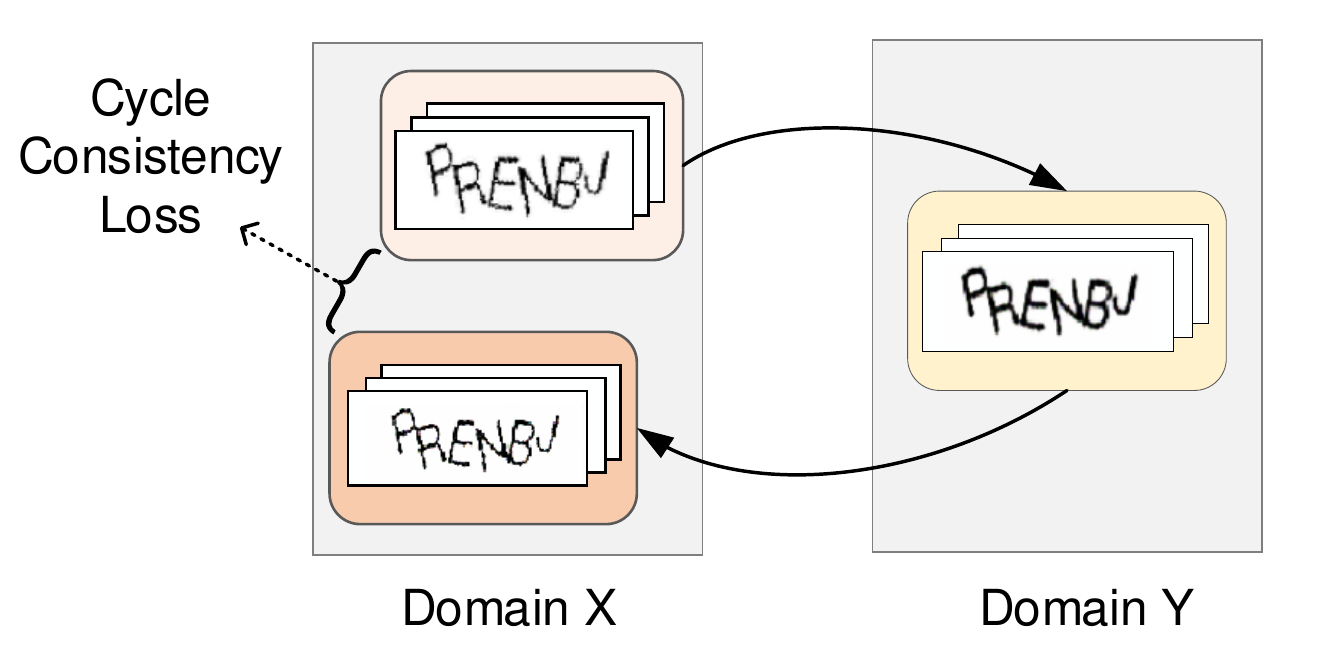}}
\caption{The training process of our CAPTCHA synthesizer: (a) The generator \(G\) extracts features from domain \(X\) and converts the extracted features into the features of domain \(Y\), and generates synthetic images. These synthetic images are fed into Generator \(F\) to obtain the reconstructed images \({F{ \left( {G{ \left( {x} \right) }} \right) }}\), where \({F{ \left( {G{ \left( {x} \right) }} \right) } \approx x}\). The discriminator \(D_{Y}\) needs to correctly distinguish the synthetic images from the real-world ones. (b) To ensure that the CAPTCHA samples in domain \(X\) can be reconstructed back to their source domain after they are converted to the target domain \(Y\), it is necessary to ensure the consistency of the conversion effect through the cycle-consistency loss.}
\label{fig:4}
\vspace{-10pt}
\end{figure*}

\par To successfully deceive the discriminators \(D_{X}\) and \(D_{Y}\), the generators \(G\) and \(F\) need to learn how to modify the input images at the pixel level so as to generate images similar to the target ones. The learning process is shown in Figure \ref{fig:4a}. The generator \(G\) encodes the CAPTCHA images \(x\) from the generation domain \(X\), extracts features from images \(x\) by using CNN, and compresses them into 256 feature vectors of size \( 8\times8 \). Then, the feature vectors of the image in the generation domain \(X\) are transformed into the feature vectors in the target domain \(Y\), and the low-dimensional features are recovered from the feature vectors using the upsampling operation to generate the synthetic CAPTCHA images \({G{ \left( {x} \right) }}\). During the upsampling process, the Transptransposed Convolution in reference \cite{zhu2017unpaired} was replaced as Sub-pixel Convolution \cite{shi2016real}, which effectively avoided Checkerboard Artifacts and improved the image quality after conversion. The generator \(F\) restores the images \({G{ \left( {x} \right) }}\) to the source domain \(X\) according to the same steps described above and generates the reconstructed images \({F{ \left( {G{ \left( {x} \right) }} \right) }}\), where \({F{ \left( {G{ \left( {x} \right) }} \right) } \approx x}\). Discriminator \(D_{y}\) is a CNN model, which extracts features through multilayer networks to distinguish whether the input images are the real-world images \(y\) or the synthetic images \(x\).
\par We define the following adversarial losses to measure the learning effect of discriminators and generators:
\begin{equation}
\begin{split}
L_{GAN}(G,D_{Y},X,Y)&=E_{y\sim P_{data}(y)}[logD_{Y}(y)] \\
&+E_{x\sim P_{data}(x)}[log(1 - D_{Y}(G(x)))]
\label{eq:adv-loss}
\end{split}
\end{equation} where the generator \(G\) tries to generate images \(G{ \left( {x} \right) }\) that look sufficiently similar to images from target domain \(Y\), while the discriminator \(D_{y}\) aims to distinguish between the synthetic images \(G{ \left( {x} \right) }\) and real-world ones \(y\). \(G\) aims to minimize the difference between synthetic images \(G{ \left( {x} \right) }\) and real-world images \(y\) against an adversarial \(D_{y}\) tries to maximize it. Therefore, the training objective defined as follows:

\begin{equation}
\setlength{\abovedisplayskip}{-5pt} 
\begin{split}
\mathop{min}\limits_{G}\mathop{max}\limits_{D_{y}}L_{GAN}(G,D_{Y},X,Y)\label{eq:training-objective}
\end{split}
\end{equation}

\par At the same time, we also define similar adversarial loss \(L_{GAN}(F,D_{X},Y,X)\) and training objective \(\mathop{min}\limits_{F}\mathop{max}\limits_{D_{x}}L_{GAN}(F,D_{X},Y,X)\) for generator \(F\) and discriminator \(D_{x}\).

\par Since there are many mapping paths between the source domain and the target domain. It means that the mapping function may map all samples in the source domain to one sample in the target domain, making adversarial losses invalid. Therefore, to reduce the possibility of such mapping and make reconstructed images consistent with source images in structure, it is necessary to keep the cyclic consistency in the process of image conversion and reconstruction (as shown in Figure \ref{fig:4b} ) to make \(F(G(x))\approx x\) and \(G(F(y))\approx y\). The formula is defined as follows:

\begin{equation}
\setlength{\abovedisplayskip}{-5pt} 
\begin{split}
L_{cyc}(G,F)&=E_{y\sim P_{data}(y)}[\left\Vert F(G(y))-y \right\Vert_{1}] \\
&+E_{x\sim P_{data}(x)}[\left\Vert G(F(x))-x \right\Vert_{1}]
\label{eq:cycle-consistency}
\end{split}
\end{equation}

\par For all the experiments, the model hyperparameters followed the default values in Cycle-GAN approach \cite{zhu2017unpaired}. We use the Adam solver \cite{kingma2015adam} with a batch size of 1 to train our CAPTCHA synthesizer. All networks were trained from scratch with a learning rate of 0.0002 and the cycle-consistency weight \(\lambda \) set to 10 which controls the relative importance of two objectives. Our final training objective is defined as:

\begin{equation}
\setlength{\abovedisplayskip}{-5pt} 
\begin{split}
G^{*},F^{*}&=arg\mathop{min}\limits_{G,F}\mathop{max}\limits_{D_{X},D_{Y}}L_{GAN}(F,D_{X},Y,X) \\
&+L_{GAN}(G,D_{Y},X,Y) + \lambda L_{cyc}(G,F)
\label{eq:total-training-objective}
\end{split}
\end{equation}

\begin{figure*}[htb]
\centering
    \vspace{-25pt}
   \includegraphics[width=0.85\textwidth]{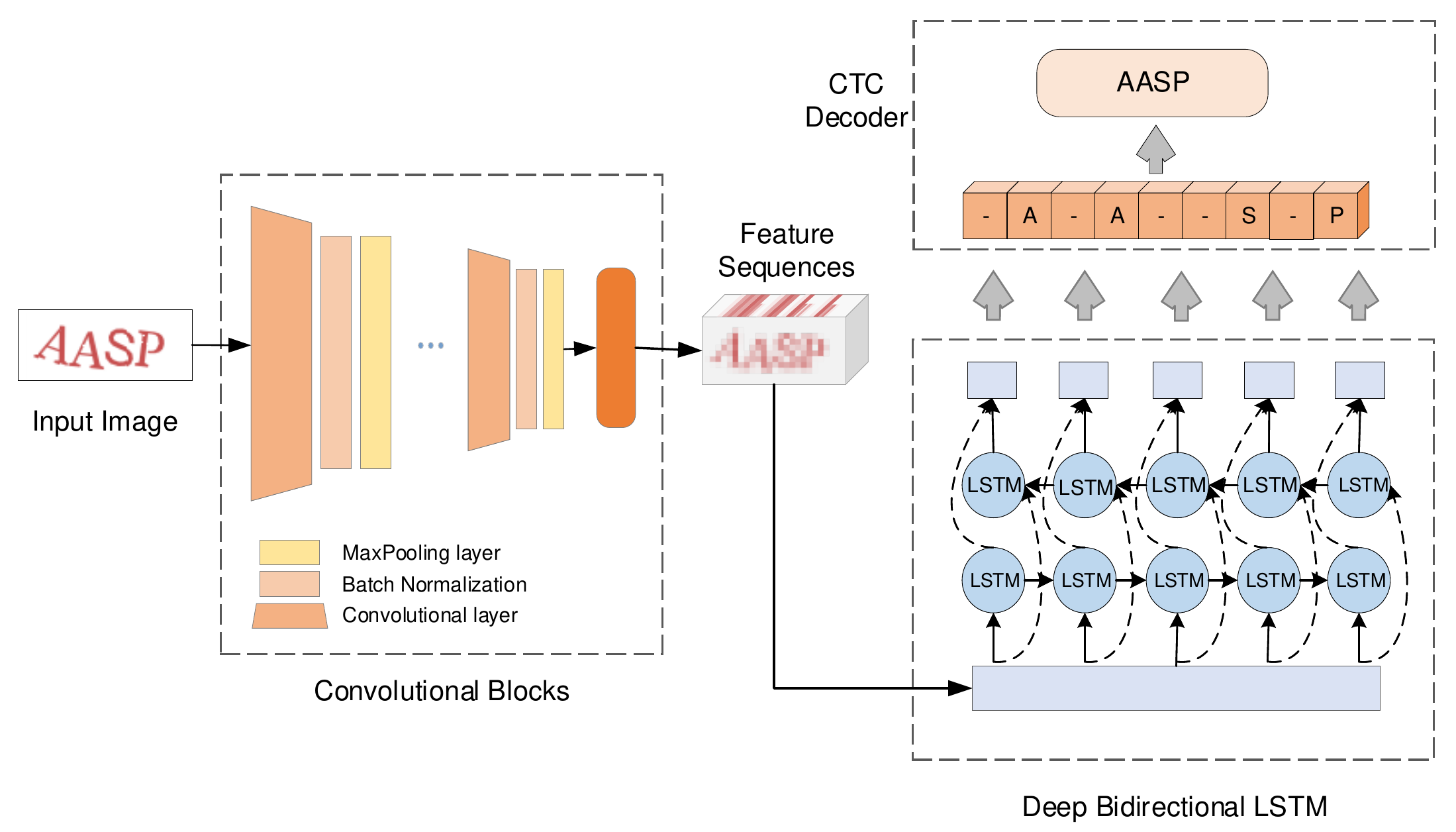}
\caption{The End-to-end CAPTCHA Breaking Network Architecture. The architecture has three components, including the convolutional blocks, the bidirectional LSTM and a CTC decoder. Firstly, the convolutional blocks consist of five convolutional layers with LeakyRelu\cite{xu2015empirical}, five batch normalizations and five max pooling layers. Secondly, the bidirectional LSTM consists of two bidirectional LSTM layers which predict a label distribution for each frame. Finally, CTC decoder translates the per-frame predictions by the recurrent layers into a label sequence.}
\label{fig:5}
\vspace{-15pt}
\end{figure*}

\par During the training process, when updating the parameters of the generator, we fixed the parameters of the discriminator. And when updating the discriminator, we fixed the parameters of the generator. For our experiments in this paper, we used middle-size data sets (contains 1,000 groups of unpaired images) and terminated the training process after 500 epochs. How to evaluate the results automatically is an open problem research field of GANs \cite{salimans2016improved,lucic2018gans}. In the case of our experiments, we evaluated the results from validation data when most synthetic CAPTCHA images are similar to real-world images. For each CAPTCHA scheme, we chose 10 CAPTCHA synthesizers that can synthesize good quality CAPTCHA images to generate some synthetic samples. We followed the perceptual study protocol\cite{isola2017image,zhang2016colorful} that participants were shown multiple groups of CAPTCHA images. Each group contained a real-world CAPTCHA image and a synthetic CAPTCHA image. And the participants were asked to click on images that they thought is real. The one which fooled most participants had been used as the final CAPTCHA synthesizer. Then the trained CAPTCHA synthesizer was used to quickly generated synthetic CAPTCHA images. In our case, we can get more than a million synthetic samples in a short time.

\subsection{End-to-End CAPTCHA Recognizer}
\par We obtained massive amounts of synthetic CAPTCHA data that are highly similar to the target CAPTCHA by using the synthesizer. Actually, there are still certain differences between the synthetic CAPTCHAs and the target ones. These differences are mainly due to the CAPTCHA generation system, because there are inevitably slight differences between the manually configured CAPTCHAs generation parameters and the target CAPTCHAs ones. The purpose of our training of GAN-based CAPTCHA synthesizer is to minimize the difference between the generated domain and the target domain. Still, this difference cannot be eliminated completely. Therefore, in our CAPTCHA breaking approach, transfer learning is used to solve the problem of model generalization. Our solution mainly consists of two steps: 1) training the basic recognizer using synthetic CAPTCHAs; 2) fine-tuning the basic recognizer obtained in step 1 using real-world samples to get the final fine-tuned recognizer. Next, we will introduce two aspects of model architecture and model training.

\par \textbf{The End-to-End CAPTCHA Breaking Network Architecture.} Our CAPTCHA recognizer is based on a CRNN whose network architecture is designed explicitly for recognizing sequence-like objects in images. It is mainly composed of CNN and RNN, and has the following advantages compared with the traditional CNN-based recognizers:
\begin{enumerate}
\item The feature information can be directly learned from the image without excessive manual preprocessing, such as segmentation, binarization, etc.
\item Insensitive to the length of the sequence objects in images, that is, in the case that the number of characters in images are not fixed, there is no need to train multiple recognizers for different character lengths.
\item The scale of the model is small, and there are fewer network parameters. The model has a quickly convergent velocity and less error.
\end{enumerate}
\begin{figure*}[htb]
\centering
    \vspace{0pt}
   \includegraphics[width=0.85\textwidth]{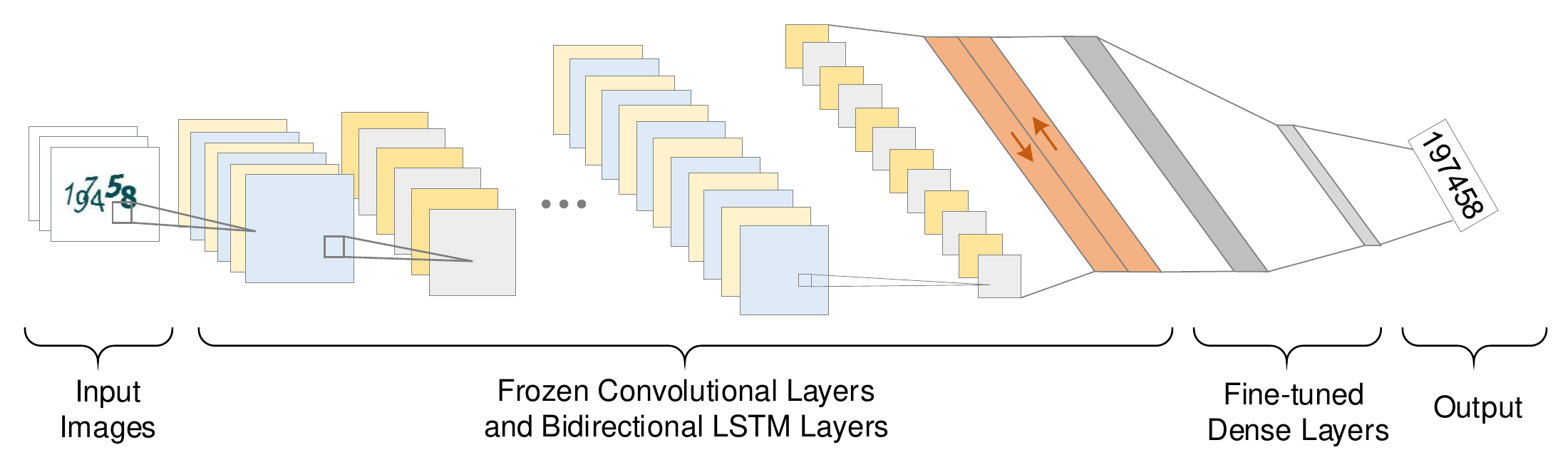}
\caption{We freeze the convolutional layers and bidirectional LSTM layers to keep the weights and then fine-tune the parameters of the top layers in the model by using the real-world CAPTCHAs.}
\label{fig:6}
\vspace{-10pt}
\end{figure*}

\begin{figure*}[htb]
\centering
    \vspace{-25pt}
   \includegraphics[width=0.85\textwidth]{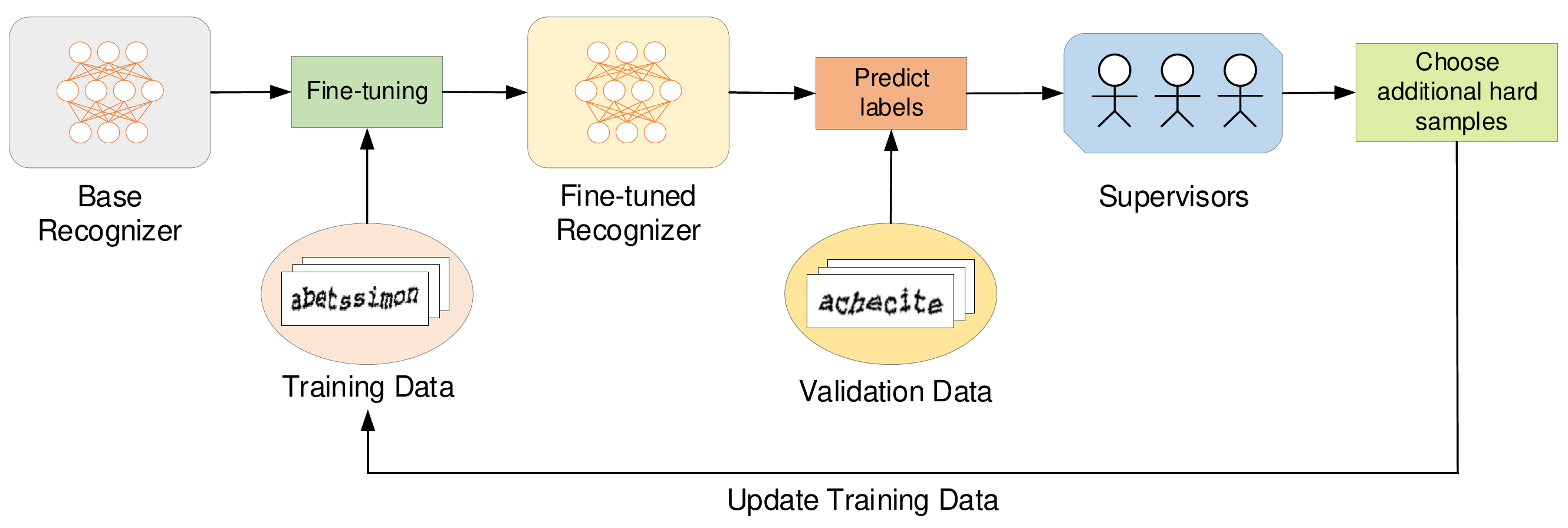}
\caption{The CAPTCHA recognizer fine-tuning process based on active learning. After several rounds of data screening and model training, the fine-tuned recognizer can achieve excellent results on the test set.}
\label{fig:7}
\vspace{-10pt}
\end{figure*}

\begin{table*}[htb]
    \vspace{5pt}
\centering
    \caption{THE ANTI-RECOGNITION MECHANISMS OF TARGET CAPTCHA SCHEMES}
    \label{tab:2}
    \renewcommand{\arraystretch}{1} 
    \renewcommand\tabcolsep{5pt} 
    \begin{tabular}{cccccccccc}
    \toprule[0.75pt]
    \textbf{Website}    & \textbf{Sample} & \textbf{\makecell[c]{Rotation \\ And Distortion}} & \textbf{Hollow} & \textbf{Overlapping} & \textbf{\makecell[c]{Noise\\ arc}} & \textbf{\makecell[c]{Variable\\ Length}} & \textbf{Multi-Fonts} & \textbf{\makecell[c]{Background\\ Interference}} & \textbf{\makecell[c]{Two-layer\\ Structure}} \\
    \midrule[0.5pt]
    Baidu(1) & \includegraphics[width=0.1\textwidth]{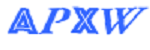} & Yes & Yes &  &  &  & Yes &  &  \\
    Baidu(2) & \includegraphics[width=0.1\textwidth]{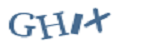} & Yes & Yes &  &  &  & Yes &  &  \\
    Baidu(3) & \includegraphics[width=0.1\textwidth]{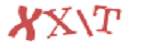} & Yes & Yes &  &  &  & Yes &  &  \\
    Amazon & \includegraphics[width=0.1\textwidth]{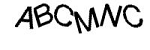} & Yes &  &  &  &  &  &  &  \\
    Wikipedia & \includegraphics[width=0.1\textwidth]{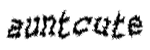} & Yes &  &  &  & Yes &  &  &  \\
    Tencent(1) & \includegraphics[width=0.1\textwidth]{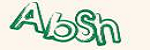} & Yes & Yes & Yes &  &  & Yes &  &  \\
    Tencent(2) & \includegraphics[width=0.1\textwidth]{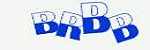} & Yes & Yes & Yes &  &  & Yes &  &  \\
    Tencent(3) & \includegraphics[width=0.1\textwidth]{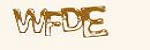} & Yes & Yes & Yes &  &  & Yes &  &  \\
    Microsoft & \includegraphics[width=0.1\textwidth]{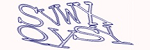} & Yes & Yes & Yes &  & Yes & Yes &  & Yes \\
    Apple & \includegraphics[width=0.1\textwidth]{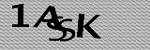} & Yes &  & Yes &  & Yes & Yes & Yes &  \\
    eBay & \includegraphics[width=0.1\textwidth]{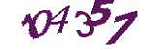} & Yes &  & Yes &  &  & Yes &  &  \\
    Sina & \includegraphics[width=0.1\textwidth]{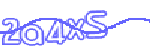} & Yes & Yes &  & Yes &  &  &  &  \\
    Sogou & \includegraphics[width=0.1\textwidth]{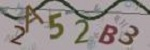} & Yes &  & Yes & Yes &  & Yes & Yes &  \\
    Weibo & \includegraphics[width=0.1\textwidth]{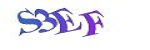} & Yes &  & Yes &  &  &  &  &  \\ \bottomrule[0.75pt]
    \end{tabular}
    \vspace{-10pt}
\end{table*}
\par In this paper, the relatively lightweight CNN5 (A network structure with five convolutional layers) and bidirectional LSTM network \cite{schuster1997bidirectional} structures are used, as shown in Figure \ref{fig:5}. It should be noted that in the selection process of convolutional networks, we had tried ResNet\cite{he2016deep} and DenseNet\cite{huang2017densely}, respectively. However, due to the larger size and more parameters of these two networks, more time is needed to train the CAPTCHA recognizers, and the breaking speed is slower than the CNN5-based recognizer. More importantly, the ResNet-based recognizer and the DenseNet-based recognizer did not significantly improve the prediction results compared to the CNN5-based recognizer, so the lightweight CNN5 was finally selected.
\par The CNN5 is mainly composed of 5 layers of repeated convolutional layers, max-pooling layers, and normalization layers, which are applied to extract continuous robust feature sequences from the input images and serve as the input of the bidirectional LSTM network. Bidirectional LSTM can effectively extract text sequence features based on convolution features and predict the label distribution of each feature vector. After that, the CTC (Connectionist Temporal Classification) decoding layer integrated all possible results of the feature sequences predicted by the LSTM to obtain the possible probability distribution of each character. It employed the label sequence with the highest probability as the final prediction string.

\par \textbf{Model Training and Fine-tuning.} For each CAPTCHA scheme, we trained a basic recognizer using the data generated by the CAPTCHA synthesizer. To obtain relatively accurate experimental results, we did a lot of pre-research experiments to search for the best training parameters. Finally, we used Adam as the optimizer and the learning rate was set to 0.0007. Besides, in the process of training the basic recognizer, the whole training process will be terminated as soon as the attack success rate on the validation set are steady and no longer fluctuates greatly. Generally, our basic recognizer will rapidly converge within 200 training epochs, and the success rate on the test set exceeds 90\%. Finally, we keep the network structure consistent and use the transfer learning method to optimize and update the network weights of the basic recognizer. During the fine-tuning process, the weight parameters of all layers except the fully connected layer on the top of the model are frozen, as shown in Figure \ref{fig:6}.

\par It is challenging to label a large number of CAPTCHA images manually that can be used to train deep learning models. In practice, it usually takes about 2 hours to label 500 real-world samples (including the time for validation). We proposed a method based on active transfer learning to reduce the need for labeled data. The main idea is to use synthetic CAPTCHAs to train a basic recognizer. Subsequently, start learning form a relatively small training set of real-world CAPTCHAs based on the basic recognizer, and then add new training samples in each subsequent round of learning. These newly added training images are not only hard samples that contain some characters cannot be correctly recognized by the current fine-tuned recognizer, but contain a large amount of useful feature information. Previous studies have showed that the purpose of breaking CAPTCHAs can be achieved with a limited number of samples\cite{ye2018yet,zi2019end}. Therefore, we set the upper limit of the training set capacity to 500 real-world samples during fine-tuning the recognizer based on active learning, and then randomly select 500 unique and labeled real-world data as the validation set.

\par Figure \ref{fig:7} shows the complete process of the active transfer learning approach. First, we fine-tune the pre-trained basic recognizer by using a small data set which contains 100 real-world CAPTCHA images. Next, we use the validation data set to test the fine-tuned recognizer and output the prediction results. By statistically comparing the predicted labels with the ground truth, we get all the characters that are misrecognized. Then selectively label 100 hard samples that contain similar characters and incrementally add them to the training set, and continue to fine-tune the recognizer. Repeat the above steps, once the training samples reach 500, stop the entire fine-tuning process. Since the training data we used in the fine-tuning process are informative samples, it can also achieve a high attack success rate with very few labeled real-world CAPTCHAs.

%
\section{Experiments}
\subsection{Data Preparation}
\par To evaluate the effectiveness of our attack method comprehensively, we have chosen 14 CAPTCHA schemes deployed by 10 websites, including Baidu, Wikipedia, Tencent, Microsoft, Apple, eBay, Weibo, Sina, Amazon, and Sogou, which are ranked in the top 50 according to the Ranking of website of Alexa from May 2019 to September 2019. We found in the survey that some websites will use the same text-based CAPTCHA scheme (such as Microsoft Live, Bing, Office, etc.). And some sites have abandoned the use of it (such as Google, LinkedIn, Alibaba, etc.) instead of using the Slider CAPTCHA scheme and the Intelligent Non-perceptive CAPTCHA scheme. Table \ref{tab:2} lists all the CAPTCHA schemes we have tested and the popular anti-recognition mechanisms adopted by these schemes. Several websites have adopted more than one CAPTCHA scheme. For example, Baidu and Tencent have deployed multiple CAPTCHA schemes to resist against CAPTCHA attacks. The main difference among these schemes is that they use a variety of different fonts.

\par The experimental process in this paper is mainly divided into three stages: 1) training of CAPTCHA synthesizer based on Cycle-GAN; 2) training of end-to-end basic recognizer; 3) fine-tuning of basic recognizer. In the synthetic model construction phase, we first generated a large number of imitation images using the CAPTCHA generation system. For each CAPTCHA scheme, because the content and structure of the CAPTCHA images are relatively simple and under the premise of without affecting the result, we only randomly chose 1,000 imitation CAPTCHAs and 1,000 real-world CAPTCHAs as training sets to minimize the time required for training the synthesizer(All real-world CAPTCHAs are collected from the target site by our web crawlers from May 2019 to September 2019). During the training phase of the end-to-end basic recognizer, we used the CAPTCHA synthesizer to generate a massive number of synthetic CAPTCHAs as training sets based on the complexity of the target CAPTCHA scheme. Then randomly selected 5,000 synthetic samples as test set and other 5,000 samples as validation set. In the phase of model fine-tuning, we adopted the fine-tuning method based on active learning. We set the maximum size of the training set for fine-tuning to 500 real-world CAPTCHA samples, and then randomly selected another 1,000 and 500 real-world ones that did not in training set as test set and validation set.
\begin{figure*}[htb]
\centering
    \vspace{-25pt}
   \subfigure[]{\label{fig:8a}\includegraphics[width=0.32\textwidth]{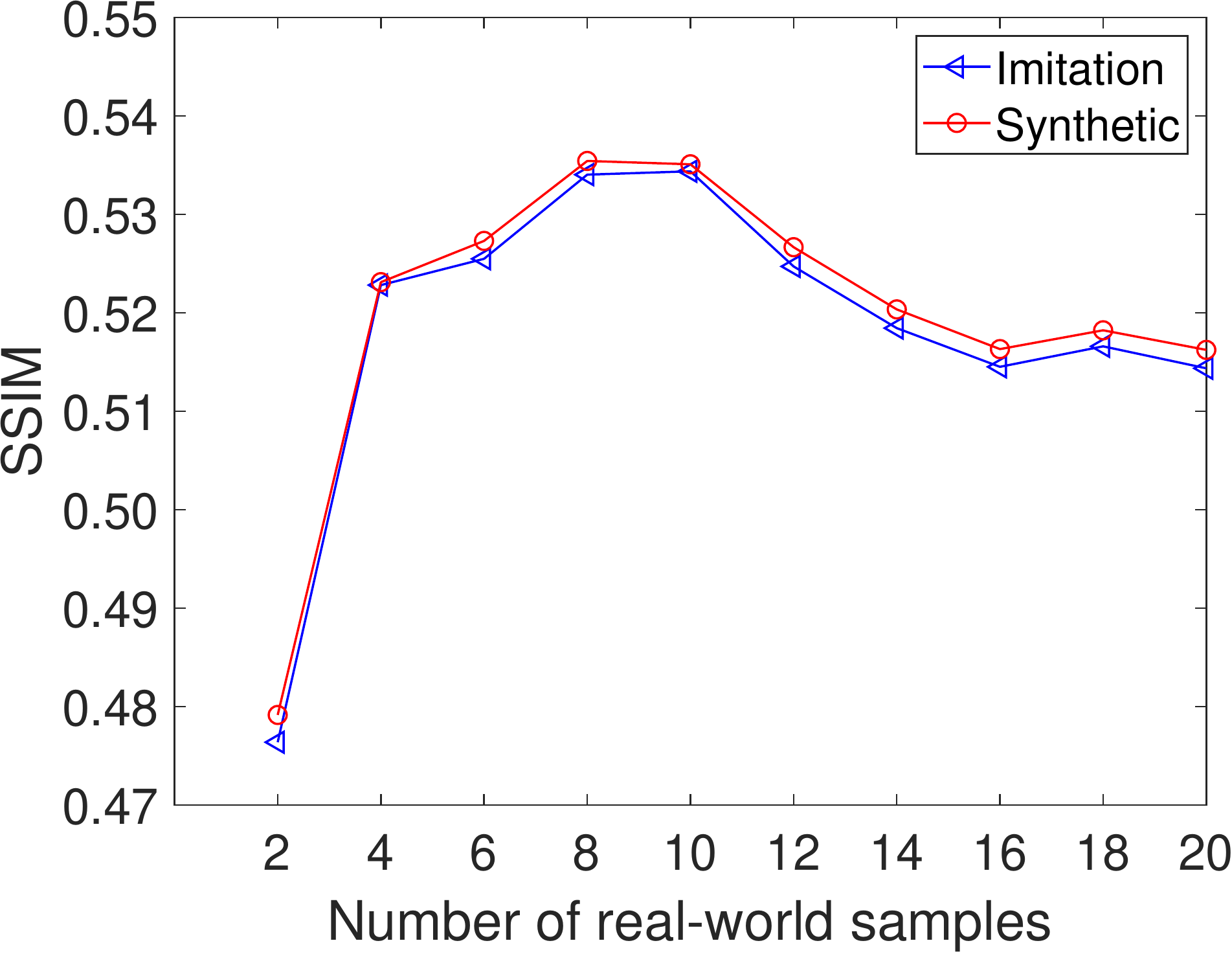}} \subfigure[]{\label{fig:8b}\includegraphics[width=0.32\textwidth]{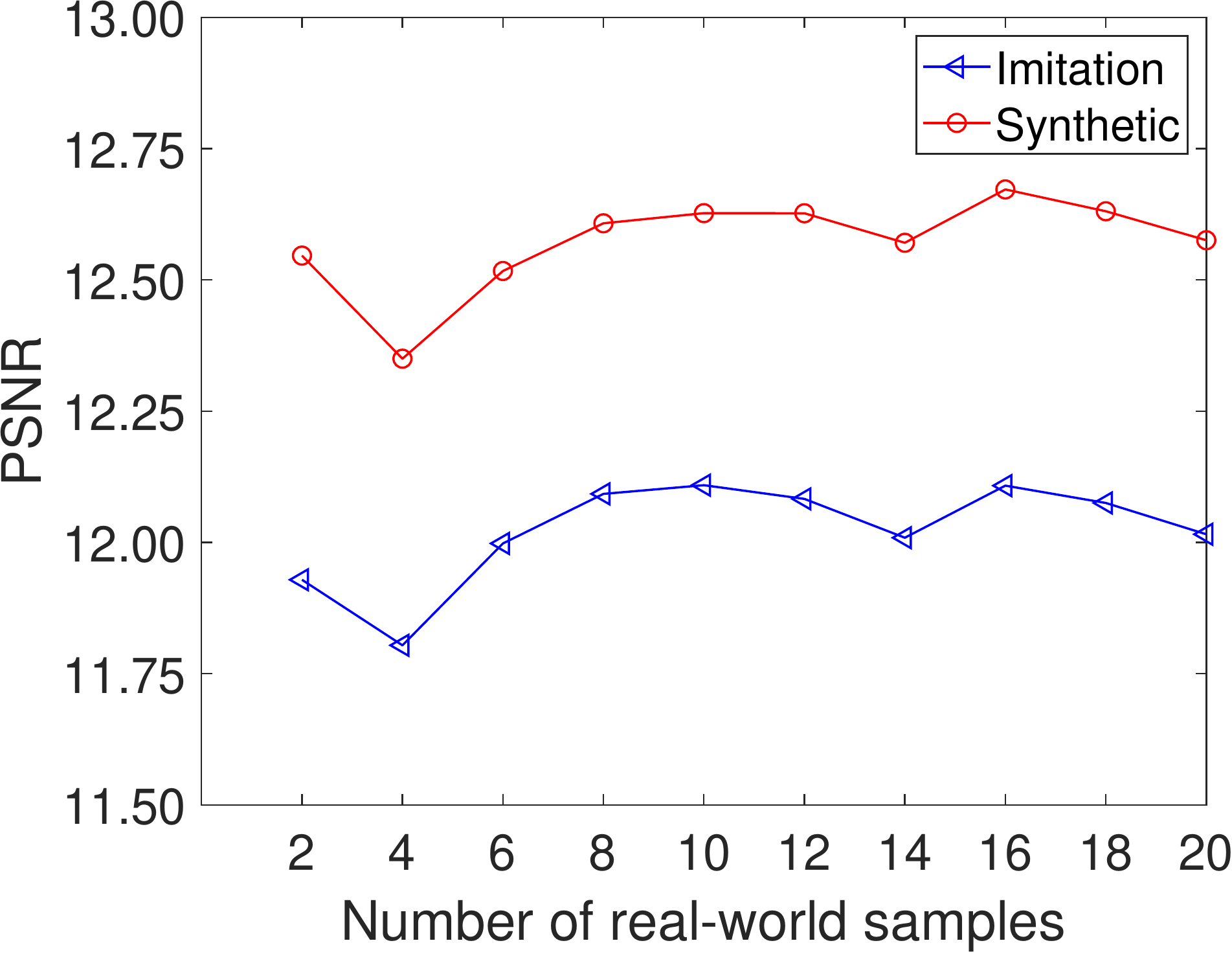}} \subfigure[]{\label{fig:8c}\includegraphics[width=0.32\textwidth]{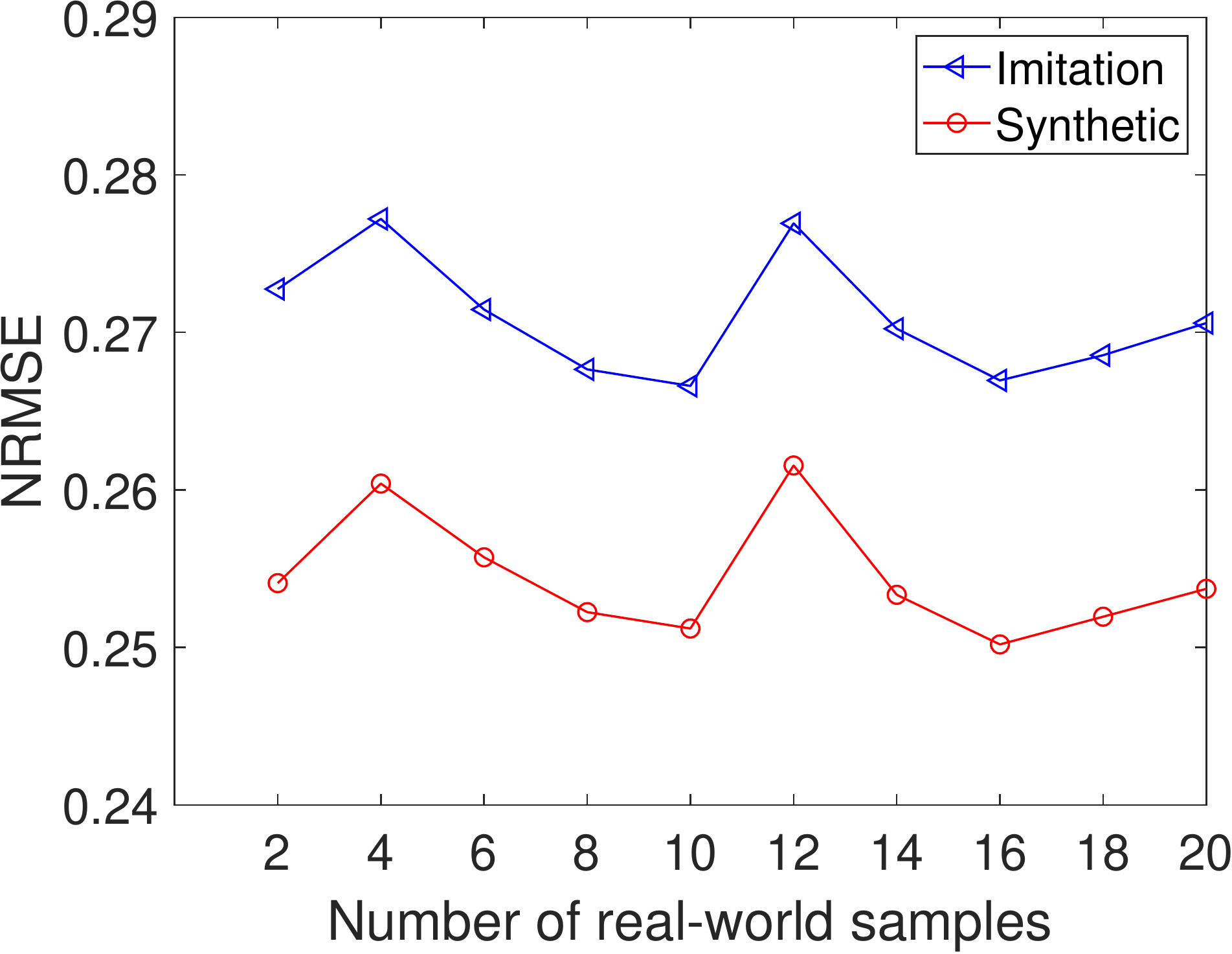}}

    \subfigure[]{\label{fig:8d}\includegraphics[width=0.32\textwidth]{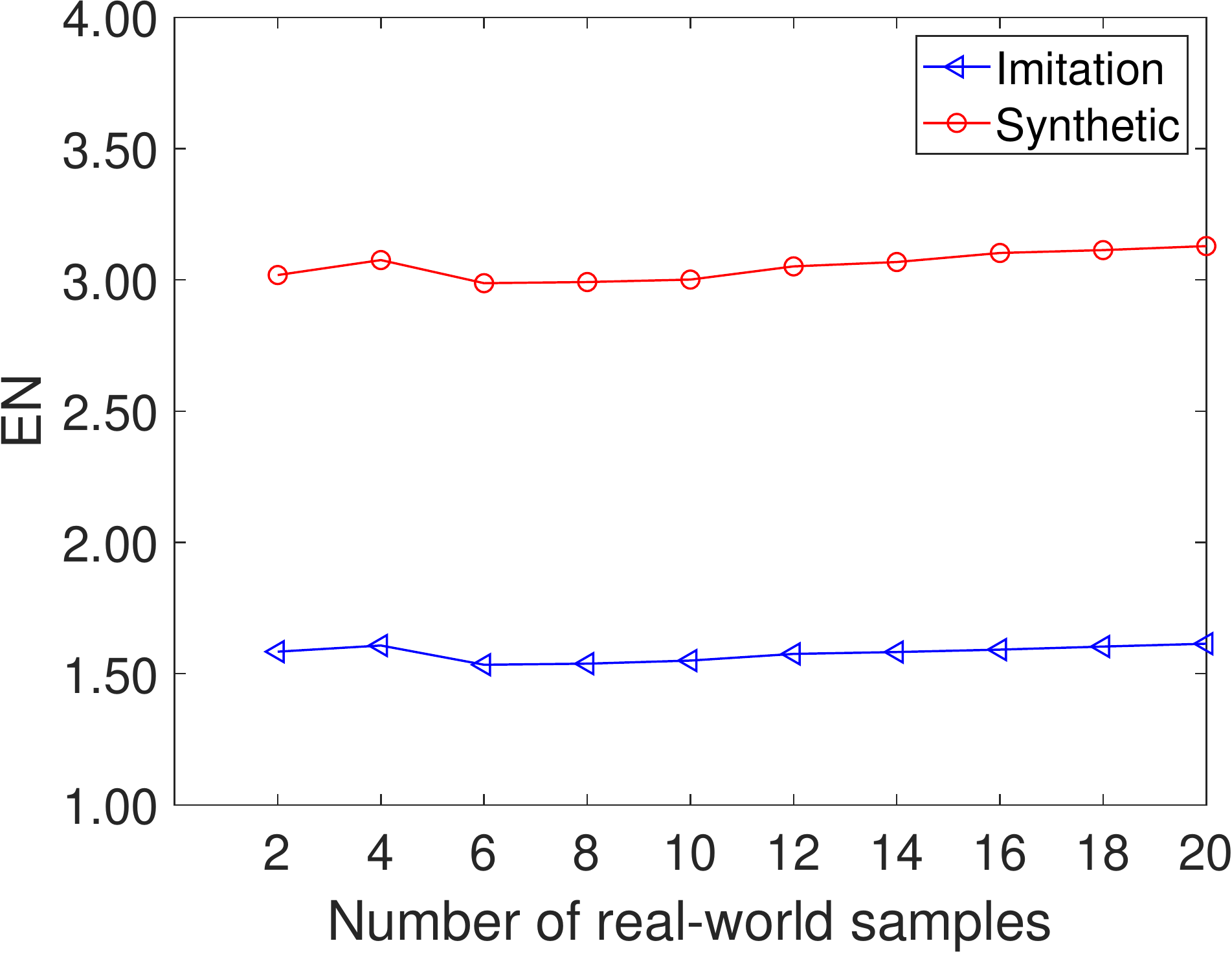}} \subfigure[]{\label{fig:8e}\includegraphics[width=0.32\textwidth]{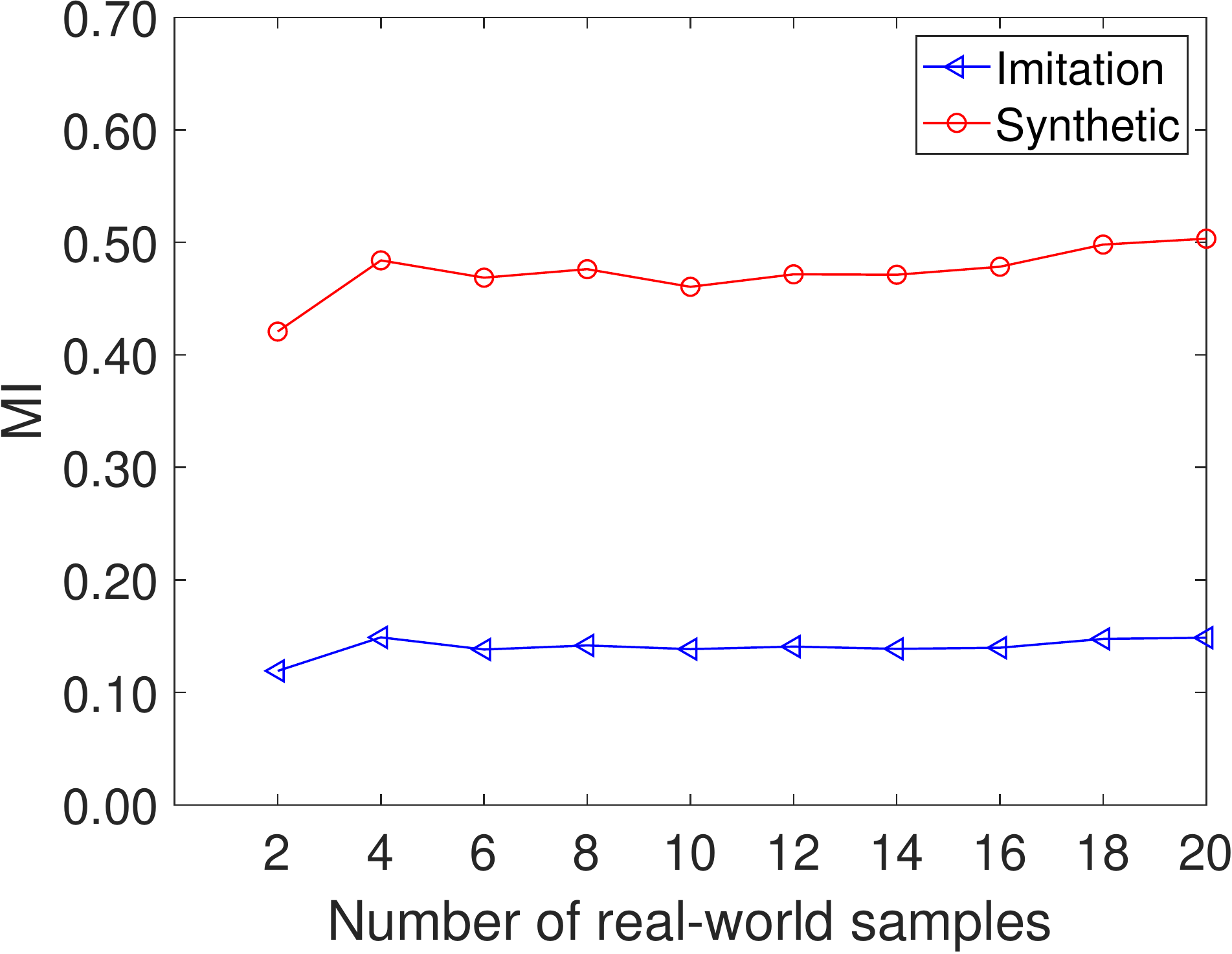}} \subfigure[]{\label{fig:8f}\includegraphics[width=0.32\textwidth]{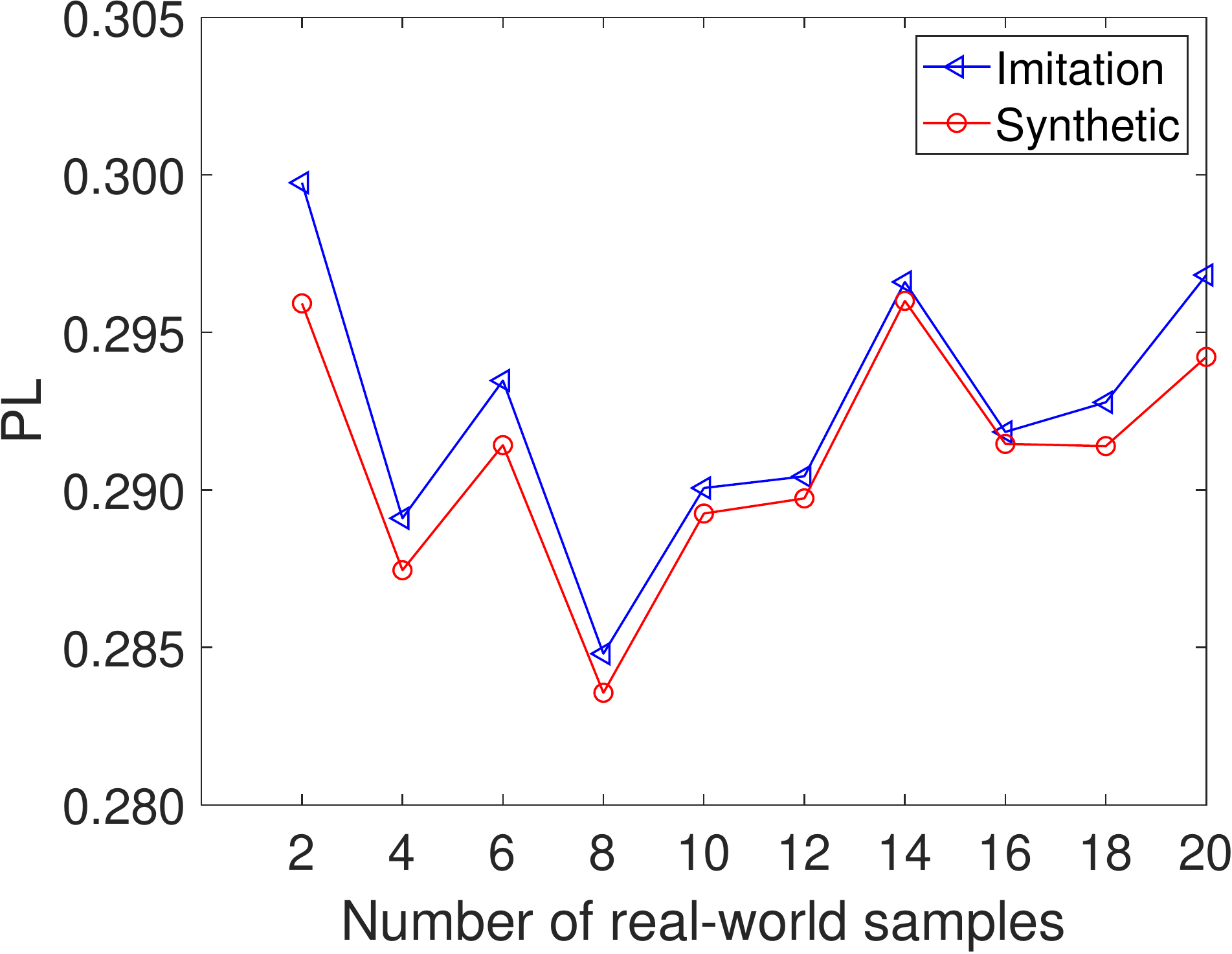}}

\caption{Quantitative comparisons. A quantitative comparison across six metrics for Baidu CAPTCHA scheme.}
\label{fig:8}
\vspace{-10pt}
\end{figure*}

\begin{table}[htb]
    \vspace{-10pt}
\centering
    \caption{THE REAL-WORLD CAPTCHAS, THE IMITATION CAPTCHAS AND THE SYNTHETIC CAPTCHAS}
    \label{tab:3}
    \setlength{\tabcolsep}{12pt}  
    \renewcommand{\arraystretch}{1} 
    \begin{tabular}{cccc}
    \toprule[0.75pt]
    \textbf{Website} & \textbf{\makecell[c]{Imitation \\ Sample}} & \textbf{\makecell[c]{Synthetic\\ Sample}} & \textbf{\makecell[c]{Real-World \\ Sample}} \\
    \midrule[0.5pt]
    Amazon & \includegraphics[width=0.08\textwidth]{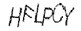} & \includegraphics[width=0.08\textwidth]{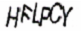} & \includegraphics[width=0.08\textwidth]{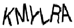}  \\
    Baidu & \includegraphics[width=0.08\textwidth]{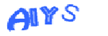} & \includegraphics[width=0.08\textwidth]{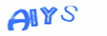} & \includegraphics[width=0.08\textwidth]{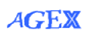}  \\
    Wikipedia & \includegraphics[width=0.08\textwidth]{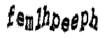} & \includegraphics[width=0.08\textwidth]{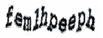} & \includegraphics[width=0.08\textwidth]{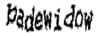}  \\
    Tencent & \includegraphics[width=0.08\textwidth]{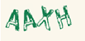} & \includegraphics[width=0.08\textwidth]{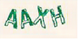} & \includegraphics[width=0.08\textwidth]{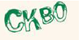}  \\
    Sina & \includegraphics[width=0.08\textwidth]{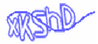} & \includegraphics[width=0.08\textwidth]{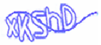} & \includegraphics[width=0.08\textwidth]{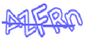}  \\
    Apple & \includegraphics[width=0.08\textwidth]{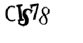} & \includegraphics[width=0.08\textwidth]{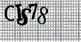} & \includegraphics[width=0.08\textwidth]{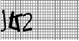}  \\
    Microsoft & \includegraphics[width=0.08\textwidth]{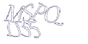} & \includegraphics[width=0.08\textwidth]{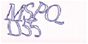} & \includegraphics[width=0.08\textwidth]{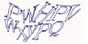}  \\
    \bottomrule[0.75pt]
    \end{tabular}
    \vspace{-10pt}
\end{table}

\subsection{Synthetic CAPTCHA Metrics}
\par Although the images synthesized by the CAPTCHA synthesizer is more like the real-word image than the images generated by CAPTCHA generation system subjectively, as shown in Table \ref{tab:3}. But it is difficult to accurately and effectively perform quality evaluation and similarity judgment on the synthetic CAPTCHAs and the imitation ones only through subjective evaluation. Therefore, different indicators are needed for objective evaluation. We have selected six different metrics to quantitatively evaluate the quality of imitation images and synthetic images, and their similarity to real-world images, including:

\begin{enumerate}
\item \textbf{SSIM (Structural Similarity Index Measure)\cite{wang2002universal}}. The SSIM algorithm is used to model image distortion. It mainly measures the similarity between the generated images and the real-world images from the aspects of luminance, contrast, and structure. The larger the SSIM value, the higher the similarity.
\item \textbf{PSNR (Peak Signal-to-Noise Ratio)\cite{huynh2008scope}}. The PSNR algorithm obtains the quality score by directly calculating the pixel gray difference value of the corresponding positions of the generated images and the real-world images. A higher score indicates that the generated image has less distortion and higher quality.
\item \textbf{NRMSE (Normalized Root-Mean-Square Error)\cite{guizar2008efficient}}. NRMSE directly measures the distance between the generated images and the real-world images in terms of pixels. The smaller the value, the closer the two images are and the more similar they are.
\item \textbf{EN (Entropy)\cite{roberts2008assessment}}. EN is used to measure the amount of information contained in the generated image. The larger the entropy, the more information it contains.
\item \textbf{MI (Mutual Information)\cite{maes1997multimodality}}. MI is used to measure the gray statistical correlation or information redundancy between the generated images and the real-world images. The larger the mutual information value between the two images, the stronger the correlation between the features contained in the images.
\item \textbf{PL (Perceptual Loss)\cite{zhang2018unreasonable}}. PL is used to evaluate the similarity of semantic features between the generated images and the real-world images. It uses a pre-trained model to extract various feature information contained in the generated image and the real-world images and then calculates the distance between the features of the two pictures. The smaller the distance, the stronger the similarity in semantic features between the generated image and the real-world one.
\end{enumerate}

\par To prove that the synthetic images are closer to the distribution of the real-world images than the original imitation images, we performed corresponding quality evaluations of the generated images on the Baidu CAPTCHA scheme and comparison with the real-world CAPTCHAs. For each evaluation method, we set up 10 groups of experiments. The first set of experiments used a total of 6 samples, including two randomly selected real-world CAPTCHA samples, two randomly selected imitation samples and their corresponding synthetic samples processed by the synthesizer. For each evaluation metric, we calculated the evaluation results between the imitation samples and the real-world samples, and the synthetic samples and the real-world samples, respectively. Following this experimental method, each subsequent set of experiments adds two additional real-world samples, two imitation samples, and two corresponding synthetic samples to the previous group. We calculate the average of the results of each group and use the mean as the final experimental results of that group. Figure \ref{fig:8} shows the evaluation results on the six metrics directly.

\begin{table}[htb]
    \vspace{0pt}
\centering
    \caption{ATTACK RESULTS WITH SYNTHETIC SAMPLES ON DIFFERENT CAPTCHA SCHEMES}
    \label{tab:4}
    \setlength{\tabcolsep}{2pt}  
    \renewcommand{\arraystretch}{1.1} 
    \begin{tabular}{ccccccc}
        \toprule[0.75pt]
        \multirow{2}{*}{\textbf{Website}} & \multicolumn{3}{c}{\textbf{Success Rate}} & \multirow{2}{*}{\textbf{Epochs}} & \multirow{2}{*}{\textbf{Speeds(ms)}} & \multirow{2}{*}{\textbf{\makecell[c]{Number\\of\\Samples}}} \\ \cmidrule(lr){2-4}
         & \multicolumn{1}{c}{\textbf{Our Attack}} & \multicolumn{1}{c}{\textbf{Ref.\cite{ye2018yet}}} & \multicolumn{1}{c}{\textbf{Ref.\cite{zi2019end}}} &  &  &  \\ \midrule[0.5pt]
        Sina & 85.0\% & 40.6\% & 17.6\% & 74 & 16 & 20,000 \\
        Tencent & 36.1\% & - & - &19 & 17 & 50,000 \\
        Amazon & 88.4\% & - & - &59 & 16 & 40,000 \\
        Baidu & 80.7\% & - & - &46 & 17 & 40,000 \\
        Wikipedia & 26.6\% & 7.0\% & - &25 & 18 & 120,000 \\
        eBay & 74.3\% & 52.0\% & - &33 & 17 & 40,000 \\
        Apple & 87.7\% & - & 85.7\% &46 & 18 & 80,000 \\
        Weibo & 79.8\% & 4.7\% & - &33 & 16 & 100,000 \\
        Sogou & 71.7\% & - & - &36 & 17 & 100,000 \\
        \makecell[c]{Microsoft\\(two-layer\\scheme)} & 22.4\% & - & - &35 & 18 & 300,000 \\
        \bottomrule[0.75pt]
    \end{tabular}
    \vspace{-5pt}
\end{table}

\par The experimental results show that the synthetic images processed by the synthesizer achieves the best results in the six evaluation indexes compared with the original imitation images. The largest SSIM and the smallest NRMSE show that our synthetic images are more similar in structure and content to the target images. The largest PSNR indicates that the synthetic images have less distortion and higher image quality than the imitation images. The maximized EN and MI show that our synthetic images contain more abundant feature information, and the feature correlation between the synthetic CAPTCHAs and the real-world ones is stronger. Minimized PL means that the semantic features carried in the synthetic images are more consistent with the feature distribution of the target images. From the results, it can be shown that the CAPTCHAs processed by the synthesizer has better quality than the original imitation CAPTCHAs and are closer to the distribution of the real-world CAPTCHAs, which is more beneficial to subsequent attacks.

\subsection{Attack with synthetic samples on different CAPTCHA schemes}

\par In order to prove that we can use the synthetic CAPTCHAs to attack the real-world CAPTCHA schemes effectively, we deployed our attack on 14 CAPTCHA schemes of 10 popular websites. For each CAPTCHA scheme, we imitated the characteristics of it and set the corresponding security feature parameters, which were loaded by the CAPTCHA generation system to generate a large number of imitation samples. These imitation samples are then fed into the CAPTCHA synthesizer to produce the corresponding processed synthetic samples. At the same time, according to the complexity of the anti-recognition mechanism of the CAPTCHA scheme used by each website, we used different numbers of synthetic samples as training data (the specific size of the training set is given in Table \ref{tab:4}) and then obtained a basic recognizer. Finally, we collected 1,000 real-world CAPTCHA images from each target website and manually labeled corresponding ground-truth to test the attack effect of the basic recognizer on real-world CAPTCHAs. Notably, for some sites, such as Baidu and Tencent, more than one kind of CAPTCHA scheme is deployed, including a variety of anti-recognition mechanisms (listed separately in Table \ref{tab:4}). Still, in the experiments, we did not do differentiate but directly mixed various schemes of the same website to attack as one scheme. Besides, our experimental platform is a server running  Centos 7 with version 3.10 of kernel, which is equipped with a 2.30GHz Intel Xeon Gold 5118 CPU, an NVIDIA RTX2080Ti GPU, and a memory size of 128GB. All model training and testing processes are completed on this server, and all models are built by using the TensorFlow 2.0 deep learning framework.

\par Table \ref{tab:4} summarizes the results of attacks on various websites using synthetic CAPTCHAs. As can be seen from the Table \ref{tab:4}, our method has a higher attack success rate in most CAPTCHA schemes, ranging from 22.4\% to 88.4\%, compared with the previous attack methods relying on synthetic CAPTCHAs \cite{ye2018yet,zi2019end}. It shows that our proposed CAPTCHA synthesis method can produce better quality and higher similarity synthetic samples than previous methods. Furthermore, the top 5 attack success rates also prove that synthetic CAPTCHAs can be directly used to launch an effective counterfeit attack on the real-world CAPTCHAs. And it is no longer necessary to label samples manually, which significantly reduces the labor costs required for the attack. Moreover, our attack method is an end-to-end process and does not require any time-consuming segmentation or pre-processing. The attack speed is breakneck that the average attack time-consuming ranges from 16 ms to 18 ms, far exceeding the rate of human recognition and the traditional segmentation-based attack methods\cite{bursztein2011text,gao2017research}.

\begin{table}[htb]
    \vspace{0pt}
    \caption{RESULTS OF ATTACKS USING DIFFERENT NUMBER OF REAL CAPTCHA SAMPLES}
    \label{tab:5}
    \setlength{\tabcolsep}{6pt}  
    \renewcommand{\arraystretch}{1.1} 
    \begin{tabular}{@{}ccccccl@{}}
        \toprule[0.75pt]
        \textbf{Website} & \textbf{\makecell[c]{basic \\ Recognizer}} & \textbf{100} & \textbf{200} & \textbf{300} & \textbf{400} & \textbf{500} \\ \midrule[0.5pt]
        Weibo & 79.8\% & 85.3\% & 88.7\% & 90.0\% & 89.5\% & 91.0\% \\
        Tencent & 36.1\% & 61.7\% & 71.1\% & 74.4\% & 74.1\% & 75.4\% \\
        Apple & 87.7\% & 88.1\% & 89.5\% & 91.4\% & 91.7\% & 93.0\% \\
        Baidu & 80.7\% & 84.6\% & 87.6\% & 88.6\% & 90.7\% & 94.0\% \\
        Wikipedia & 26.6\% & 75.1\% & 80.5\% & 83.5\% & 86.6\% & 87.5\% \\
        Sogou & 71.7\% & 72.4\% & 75.6\% & 77.3\% & 81.7\% & 86.6\% \\
        eBay & 74.3\% & 85.1 \% & 87.1\% & 88.4\% & 91.7\% & 92.1\% \\
        Sina & 85.0\% & 90.0\% & 94.6\% & 94.6\% & 96.0\% & 97.6\% \\
        Amazon & 88.4\% & 95.1\% & 95.9\% & 96.5\% & 96.6\% & 97.3\% \\
       \makecell[c]{Microsoft\\(two-layer\\scheme)} & 22.4\% & 23.6\% & 26.7\% & 29.7\% & 32.2\% & 33.8\% \\ \bottomrule[0.75pt]
        \end{tabular}
    \vspace{0pt}
\end{table}

\par However, it can also be found from the table that, for the CAPTCHA schemes of some websites, such as Tencent, Wikipedia, and Microsoft, the breaking success rates are relatively low, respectively 36.1\%, 26.6\%, and 22.4\%. Obviously, the similarities between the synthetic CAPTCHAs and the real-world CAPTCHAs are still not good enough. Several factors lead to this result: (1) The CAPTCHA schemes of Tencent and Microsoft websites adopt a variety of fonts. Actually, it is difficult to gather the corresponding fonts, and we can only use fonts as similar as possible to replace them. Moreover, the CAPTCHA of  Microsoft website adopts the anti-recognition mechanism with a two-layer structure and distortion, which makes it difficult for the generation system and synthesizer to imitate the real-world CAPTCHAs completely. Therefore, there will be slight differences between synthetic samples and real-world samples. Although it does not affect the successful recognition of images by humans, but for deep neural networks, these subtle differences will be amplified after multiple activation layers, and will eventually cause incorrect output. (2) The length of the CAPTCHA content of Wikipedia and Microsoft websites is generally too long (it typically contains  8 to 10 characters), which means that the model needs to recognize longer sequences. Any character prediction error will make this attack task fail, thus increasing the difficulty of attack.

\subsection{Attack based on active transfer learning}
\begin{table*}[]
    \vspace{-15pt}
    \centering
    \caption{COMPARED WITH THE RESULTS OF PREVIOUS ATTACKS}
    \label{tab:6}
    \setlength{\tabcolsep}{3pt}  
    \renewcommand{\arraystretch}{1} 
    \begin{tabular}{@{}ccccccc@{}}
        \toprule
        \multirow{2}{*}{\textbf{Website}} & \multicolumn{2}{c}{\textbf{Our Attack}} & \multicolumn{2}{c}{\textbf{Ref.\cite{ye2018yet}}} & \multicolumn{2}{c}{\textbf{Ref.\cite{zi2019end}}} \\ \cmidrule(l){2-7}
         & \multicolumn{1}{c}{\textbf{Success Rate}} & \multicolumn{1}{c}{\textbf{\makecell[c]{Number of\\ Real-World Samples}}} & \multicolumn{1}{c}{\textbf{Success Rate}} & \multicolumn{1}{c}{\textbf{\makecell[c]{Number of\\ Real-World Samples}}} & \multicolumn{1}{c}{\textbf{Success Rate}} & \multicolumn{1}{c}{\textbf{\makecell[c]{Number of\\ Real-World Samples}}}\\ \midrule
        Weibo & \textbf{91.0\%} & \textbf{500} & 44.0\%& 500& 90.2\% & 8,000 \\
        Tencent & 75.4\% & 500 & - & - & - & -\\
        Apple & \textbf{88.1\%} & \textbf{100} & - & - & 83.7\% & 8,000 \\
        Baidu & \textbf{94.0\%} & \textbf{500} & - & - & 92.2\% & 2,000 \\
        Wikipedia & \textbf{87.5\%} & \textbf{500} & 78.0\% & 500 & 86.8\% & 2,000 \\
        Sogou & 86.6\% & 500 & - & - & - & - \\
        eBay & \textbf{87.1\%} & \textbf{200} & 86.6\% & 500 & - & -\\
        Sina & \textbf{90.0\%} & \textbf{100} & 52.6\% & 500 & 86.2\% & 10,000 \\
        Amazon & \textbf{95.1\%} & \textbf{100} & 79.0\% & 500 & - & -\\
        Microsoft(two-layer scheme) & 33.8\% & 500 & - & - & - & -\\
        \bottomrule[0.75pt]
    \end{tabular}
    \vspace{-10pt}
\end{table*}
\par To solve the problem of low attack success rate caused by the difference between the synthetic CAPTCHAs and the real-world CAPTCHAs, we utilize the method based on active transfer learning to train the basic recognizer to further improve the attack success rates on the target CAPTCHA schemes. Table \ref{tab:5} summarizes the attack results after fine-tuning the basic recognizer with different numbers of real-world samples. We can find from the table that with the continuous increase of real-world samples, the success rates of attacks have been improved. After fine-tuning with 100 real-world samples, for those CAPTCHA schemes that make the basic recognizer get lower success rates, such as Tencent, Wikipedia, and eBay, the attack success rates are greatly improved, with an increase ranging from 10.8\% to 48.5\%. The main reason is that the training data we selected is the hard samples that contain more useful feature information. Therefore, the model can learn from these samples and then adjust the network parameters to reduce the error caused by the synthetic CAPTCHAs, thereby better fitting the real-world CAPTCHAs distribution. When using 500 real samples, the attack success rate on 9 websites exceeded 75\%, and the attack success rate on 6 sites exceeded 90\%. It shows that these CAPTCHA schemes cannot effectively resist our attack. At the same time, it is not difficult to find that most of the CAPTCHA schemes that have been successfully cracked adopt fewer anti-recognition mechanisms. For example, the CAPTCHA scheme of Amazon only uses rotation; the CAPTCHA scheme of Sina only employs hollow fonts and noise arcs. It is worth mentioning that although the attack success rate of Tencent CAPTCHA scheme is quite low (only 75.4\%). But we found in the process of collecting real-world samples that its CAPTCHA sample pool is minimal (only contains about 4,500 CAPTCHA images), which is very vulnerable to MD5 dictionary attack. Also, we can find that the attack effect on Microsoft is relatively inadequate. After fine-tuning with 500 samples, it can only achieve a success rate of 33.8\%. The main reason is that the two-layer structure scheme, multiple artistic fonts, and excessive distortion of image content, which still leaves a large gap between our synthetic CAPTCHAs and the real-world Microsoft CAPTCHAs. Only a small amount of fine-tuning training data cannot significantly reduce the difference.

\par
\begin{figure}[htb]
\centering
    \vspace{-5pt}
  \includegraphics[width=0.48\textwidth]{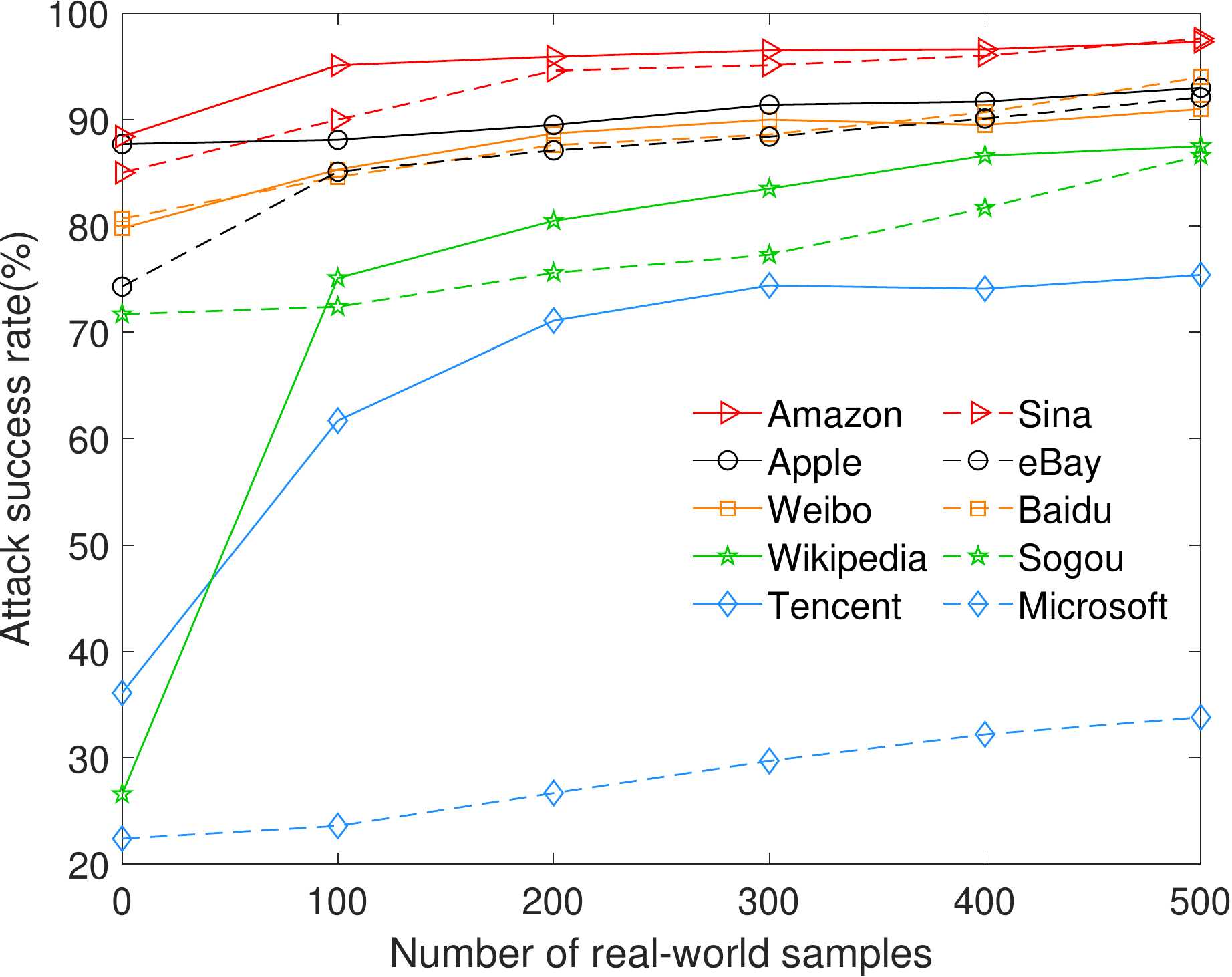}
\caption{The impact of the number of real-world CAPTCHA samples for the success rate of fine-tuned model.}
\label{fig:9}
\vspace{-5pt}
\end{figure}

 To directly demonstrate the accuracy of each CAPTCHA scheme when using different numbers of training data, we present a graph, as shown in Figure \ref{fig:9}. In general, with the increasing of real-world samples, the attack success rates also continue to increase. However, for Weibo and Tencent CAPTCHA schemes, after the training set is increased from 300 to 400, the success rate decrease slightly. The main reason is that the training set is smaller than test set, after the new samples are added, the training set generates sample imbalance problem, which makes the data distribution of the training set and the test set considerably different. As a result, the accuracy is reduced.

\par For most CAPTCHA schemes, such as Apple, eBay, Tencent, Sina, Amazon, Weibo, and Baidu, when the number of training samples used for fine-tuning exceeded 200, the model achieved a high attack success rate. With the further increase of samples, the success rate improved, but the effect was not noticeable. And the overall success rate tended to be stable. Compared with previous studies\cite{ye2018yet,zi2019end} (as shown in Table \ref{tab:6}), for each target CAPTCHA scheme, our method requires fewer real-world CAPTCHAs when achieving similar or higher success rates, ranging from 100 samples to 500 samples. The results show that our proposed method based on active transfer learning can undoubtedly improve the breaking effect while using small-scale data sets. It effectively reduces the workload of manually labeled samples, and significantly reducing the difficulty of breaking text-based CAPTCHAs.

\subsection{Effect of varied CAPTCHA security mechanisms}\label{section:4.5}
\par In this experiment, we explored the effectiveness of various popular CAPTCHA security mechanisms to provide a reference for designing more robust CAPTCHA schemes. We consider a total of 12 security schemes, including 8 single security schemes such as rotation, character overlap, image content distortion, interference lines, etc, and 4 combined security schemes, as shown in Table \ref{tab:7}. For each scenario, we used the CAPTCHA generation system to generate 40,000 samples as the training set, 4,000 samples as the validation set, and 4,000 samples as the test set. During training, when the accuracy of the recognizer on the validation set has stabilized and no longer changes, we terminate the training process and test the CAPTCHA recognizer on the test set and get the predicted results, as shown in Table \ref{tab:7}.

\begin{table*}[htb]
    \vspace{-15pt}
\centering
    \caption{TWELVE KINDS OF CAPTCHA SECURITY MECHANISMS}
    \label{tab:7}
    \renewcommand{\arraystretch}{1} 
    \renewcommand\tabcolsep{3pt} 
    \begin{tabular}{cccccccccccc}
    \toprule[0.75pt]
    \textbf{No} & \textbf{Sample} & \textbf{\makecell[c]{Rotation}} & \textbf{Overlapping} & \textbf{Distortion} & \textbf{Multi-Fonts} & \textbf{\makecell[c]{Noise\\arc}} & \textbf{\makecell[c]{Variable\\ Length}}  & \textbf{\makecell[c]{Background\\ Interference}} & \textbf{\makecell[c]{Two-layer\\ Structure}} & \textbf{\makecell[c]{Ground\\Truth}} & \textbf{\makecell[c]{Success\\Rate}}\\
    \midrule[0.5pt]
    1 & \includegraphics[width=0.1\textwidth]{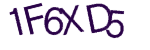} & Yes &  &  &  &  &  &  &  & 1F6XD5 & 97.4\% \\
    2 & \includegraphics[width=0.1\textwidth]{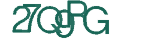} &  & Yes &  &  &  &  &  &  & 27Q9PG & 95.1\% \\
    3 & \includegraphics[width=0.1\textwidth]{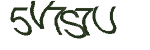} &  &  & Yes &  &  &  &  &  & 5V7S7U & 96.1\% \\
    4 & \includegraphics[width=0.1\textwidth]{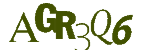} &  &  &  & Yes &  &  &  &  & AGR3Q6 & 94.9\% \\
    5 & \includegraphics[width=0.1\textwidth]{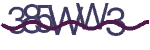} &  &  &  &  & Yes &  &  &  & 385WW3 & 97.5\% \\
    6 & \includegraphics[width=0.1\textwidth]{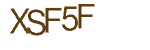} &  &  &  &  &  & Yes &  &  & XSF5F & 96.0\% \\
    7 & \includegraphics[width=0.1\textwidth]{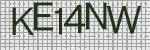} &  &  &  &  &  &  & Yes &  & KE14NW & 96.6\% \\
    8 & \includegraphics[width=0.1\textwidth]{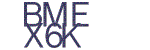} &  &  &  &  &  &  &  & Yes & BMEX6K & 94.4\% \\
    9 & \includegraphics[width=0.1\textwidth]{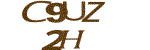} &  & Yes &  & Yes &  &  &  & Yes & C9UZ2H & 80.0\% \\
    10 & \includegraphics[width=0.1\textwidth]{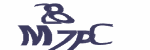} &  & Yes & Yes & Yes &  & Yes &  & Yes & 28M7PC & 62.1\% \\
    11 & \includegraphics[width=0.1\textwidth]{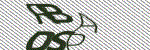} &  & Yes & Yes & Yes &  & Yes & Yes & Yes & RBADSP & 47.4\% \\
    12 & \includegraphics[width=0.1\textwidth]{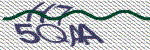} & Yes & Yes & Yes & Yes & Yes & Yes & Yes & Yes & H75QAA & 10.6\% \\
     \bottomrule[0.75pt]
    \end{tabular}
    \vspace{-10pt}
\end{table*}

\begin{figure}[htb]
\centering
    \vspace{-5pt}
  \includegraphics[width=0.45\textwidth]{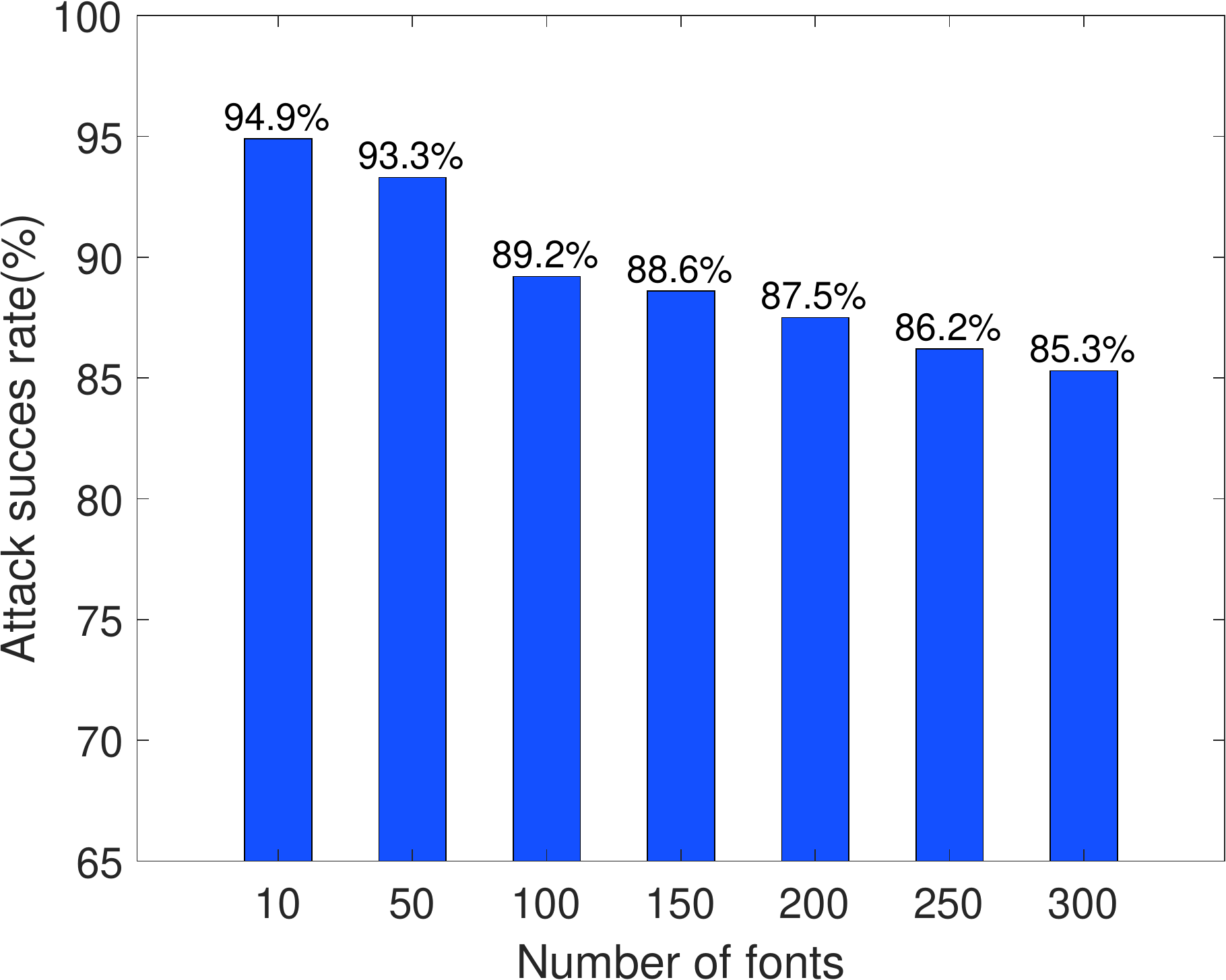}
\caption{The impact of the number of fonts for attack success rate.}
\label{fig:10}
\vspace{-5pt}
\end{figure}

\par We can find from the Table \ref{tab:7} that all the CAPTCHA schemes using these 12 security mechanisms have been cracked. Among the first eight CAPTCHA schemes with a single security mechanism, our method can achieve a success rate higher than 94\%, which also indicates that the CAPTCHA schemes using only one security mechanism is not secure. And it cannot effectively resist automated attacks. Although the CAPTCHA with only one security mechanism had low security, but the attack success rate dropped sharply from 97.5\% to 10.6\% as more security mechanisms were added. For example, scheme No.12 adds only two additional security mechanisms based on scheme No.11, but the attack success rate decreased from 47.4\% to 10.6\%, with a drop of 36.8\%. Obviously, combining more security mechanisms plays an essential role in designing more robust CAPTCHA schemes.

\par Besides, we also noticed that among the first eight schemes with a single security mechanism, the CAPTCHA scheme using multi-fonts security mechanism is slightly more secure than the other six schemes. Therefore, we designed an additional experiment to verify whether the scheme using only multi-fonts security mechanism can achieve the similar effect of the scheme using combined security mechanisms. The experimental results are shown in Figure \ref{fig:10}, where the abscissa represents the number of used font types. Although the success rate is gradually decreasing with the increase in the types of used fonts, but the resistance effect is not very prominent. It is not better than the scheme No. 9, which uses three security mechanisms. Therefore, if only multi-fonts security mechanism is used in the design of CAPTCHA schemes, even if a large number of fonts are used, the attack of our method cannot be effectively resisted. And collecting a massive amount of fonts is a very tedious task, which requires a volume of workforce and time to check the availability of each font. 
\par As a result, it is crucial to design a CAPTCHA scheme that can resist the attack based on deep learning. We think that there are two CAPTCHA design ideas: one is to fool the CAPTCHA recognizers based on machine learning or deep learning by using adversarial examples\cite{goodfellow2015explaining,kurakin2017adversarial}. That is, by adding some disturbances that are hard to detect by human to CAPTCHA images so that the normal CAPTCHA recognizers output wrong prediction results. Another design idea is to make full use of human excellent logical reasoning, semantic understanding, and other abilities to enhance the robustness of CAPTCHA. The reason is that the advanced behaviors of creatures are currently challenging to be imitated by computer.

\section{Conclusion}
\par In this paper, we proposed an end-to-end text-based CAPTCHA breaking method based on generative adversarial networks. We trained cycle-GAN-based synthesizers to generate a great number of synthetic CAPTCHA samples, which solves the problem of lacking of training data. And we proposed a model training method based on active transfer learning, using a small amount of hard samples with more information to fine-tune the pre-trained deep learning models. We also cracked a large number of real-world CAPTCHAs that are deployed by the top 50 websites as ranked by Alexa.com. The success rates ranged from 33.8\% to 97.6\%. Compared with previous studies, our approach requires less labeled real-world data when reaching the same or higher attack success rates, ranging from 100 to 500 samples. And our method has better versatility. For a new text-based CAPTCHA scheme, we can quickly break it without investing a lot of labor and time.

\par At the same time, We analyzed the security of 12 popular anti-recognition mechanisms. The experimental results demonstrate that more security mechanisms can improve the safety of the CAPTCHAs. Nevertheless, once attackers obtain a massive number of training data through various methods, these security mechanisms may no longer be valid. This shows that the current text-based CAPTCHA schemes are insecure and cannot resist attacks based on machine learning and deep learning. Finally, we proposed suggestions for the future development of CAPTCHAs based on the shortcomings of the existing CAPTCHA schemes. We also hope that the work of this paper can provide a reference for security experts to design some CAPTCHA schemes with both security and availability.
%
\section*{Acknowledgment}
This work was supported by the National Science and Technology Support Programme of China under grant no. 2012BAH18B05 and the National Natural Science Foundation of China (NSFC) under grant nos. 61802271, 81602935, and U19A2081. The authors thank anonymous reviewers for their helpful comments to improve the paper.

%

\end{document}